\documentclass[letterpaper, 10 pt, journal, twoside]{IEEEtran}
\usepackage[T1]{fontenc}
\usepackage{makecell}
\usepackage{array}
\usepackage{subfigure}
\usepackage{multirow}
\usepackage{threeparttable}
\usepackage{algorithm}
\usepackage{algpseudocode}
\usepackage{makecell}
\usepackage{hyperref}
\usepackage{graphics}           
\usepackage{times}              
\usepackage{amsmath}            
\usepackage{amssymb}            
\usepackage{graphicx}
\usepackage{algorithm}
\usepackage{algpseudocode}
\usepackage{booktabs}
\usepackage{color}
\definecolor{instructioncolor}{rgb}{.5,.5,.5}

\usepackage[font=small]{caption}

\def\figref#1{Fig.~\ref{#1}}
\def\tabref#1{Tab.~\ref{#1}}
\def\eqref#1{Eq.~(\ref{#1})}

\makeatletter
\usepackage{xspace}
\DeclareRobustCommand\onedot{\futurelet\@let@token\@onedot}
\def\@onedot{\ifx\@let@token.\else.\null\fi\xspace}

\def\etal{{et al}\onedot}
\makeatother

\usepackage{array}
\newcolumntype{L}[1]{>{\raggedright\let\newline\\\arraybackslash\hspace{0pt}}m{#1}}
\newcolumntype{C}[1]{>{\centering\let\newline\\\arraybackslash\hspace{0pt}}m{#1}}
\newcolumntype{R}[1]{>{\raggedleft\let\newline\\\arraybackslash\hspace{0pt}}m{#1}}

\title{\LARGE \bf Flexible Exoskeleton Control Based on Binding Alignment Strategy and Full-arm Coordination Mechanism}

\author{Chuang\,Cheng,~Xinglong\,Zhang,~Xieyuanli\,Chen,~Wei\,Dai,~Longwen\,Chen,~Daoxun\,Zhang,~Hui\,Zhang,\\~Jie\,Jiang,~Huimin\,Lu% <-this % stops a space
  \thanks{All authors are with the College of Intelligence Science and Technology, National University of Defense Technology, Changsha, China. Jie Jiang is also with the China Academy of Launch Vehicle Technology in Beijing, China.}%
  \thanks{This work was supported in part by the National Science Foundation of China under Grant 62203460, U22A2059, and 62403478, Young Elite Scientists Sponsorship Program by CAST (No. 2023QNRC001), as well as the Innovation Science Foundation of National University of Defense Technology under Grant 24-ZZCX-GZZ-11. (Corresponding author: Huimin Lu, Hui Zhang.)
  }%
}

\begin{document}
\maketitle
\markboth{}
{C. Cheng~\etal: Flexible Exoskeleton Control Based on Binding Alignment Strategy and Full-Arm Coordination Mechanism}

%%%%%%%%%%%%%%%%%%%%%%%%%%%%%%%%%%%%%%%%%%%%%%%%%%%%%%%%%%%%%%%%%%%%%%%%%%%%%%%%
\begin{abstract}
In rehabilitation, powered, and teleoperation exoskeletons, connecting the human body to the exoskeleton through binding attachments is a common configuration. However, the uncertainty of the tightness and the donning deviation of the binding attachments will affect the flexibility and comfort of the exoskeletons, especially during high-speed movement. 
To address this challenge, this paper presents a flexible exoskeleton control approach with binding alignment and full-arm coordination. Firstly, the sources of the force interaction caused by donning offsets are analyzed, based on which the interactive force data is classified into the major, assistant, coordination, and redundant component categories. Then, a binding alignment strategy~(BAS) is proposed to reduce the donning disturbances by combining different force data. Furthermore, we propose a full-arm coordination mechanism~(FCM) that focuses on two modes of arm movement intent, joint-oriented and target-oriented, to improve the flexible performance of the whole exoskeleton control during high-speed motion. In this method, we propose an algorithm to distinguish the two intentions to resolve the conflict issue of the force component. Finally, a series of experiments covering various aspects of exoskeleton performance (flexibility, adaptability, accuracy, speed, and fatigue) were conducted to demonstrate the benefits of our control framework in our full-arm exoskeleton.

\end{abstract}
\vspace{-0.4cm}
%%%%%%%%%%%%%%%%%%%
\section{Introduction}
\label{sec:intro}
Upper limb exoskeletons have various applications in rehabilitation~\cite{zimmermann2023tro,kim2017ijrr}, powered assistance~\cite{Kim2020rcim}, teleoperation~\cite{rebelo2014ram}, and other scenarios~\cite{Sun2021ral}. In addition to the design of the mechanism and system in upper limb exoskeleton research~\cite{Huang2016tsmc,kim2017ijrr,Wu2018tsmc}, another crucial research area is the control of exoskeletons~\cite{Wang2021ral,zimmermann2020iros, Li2020tsmc, Huang2019tsmc}. 
In these fields, the flexibility of the exoskeleton is an important performance~(i.e., the force required by users to drive the exoskeleton) that directly affects
the user experience and the operational feel. Especially for teleoperation exoskeletons, researchers expect that exoskeleton control can allow users to operate the exoskeleton freely, comfortably, and effortlessly, similar to the natural movement of their arms. In powered assistance and rehabilitation applications, the control objective can also serve as the baseline, which adds assistance or resistance force by adjusting the control gain. Therefore, research into fast and flexible exoskeleton control is particularly important.

\begin{figure}[t]
  \centering
  \includegraphics
  [width=0.9\linewidth]{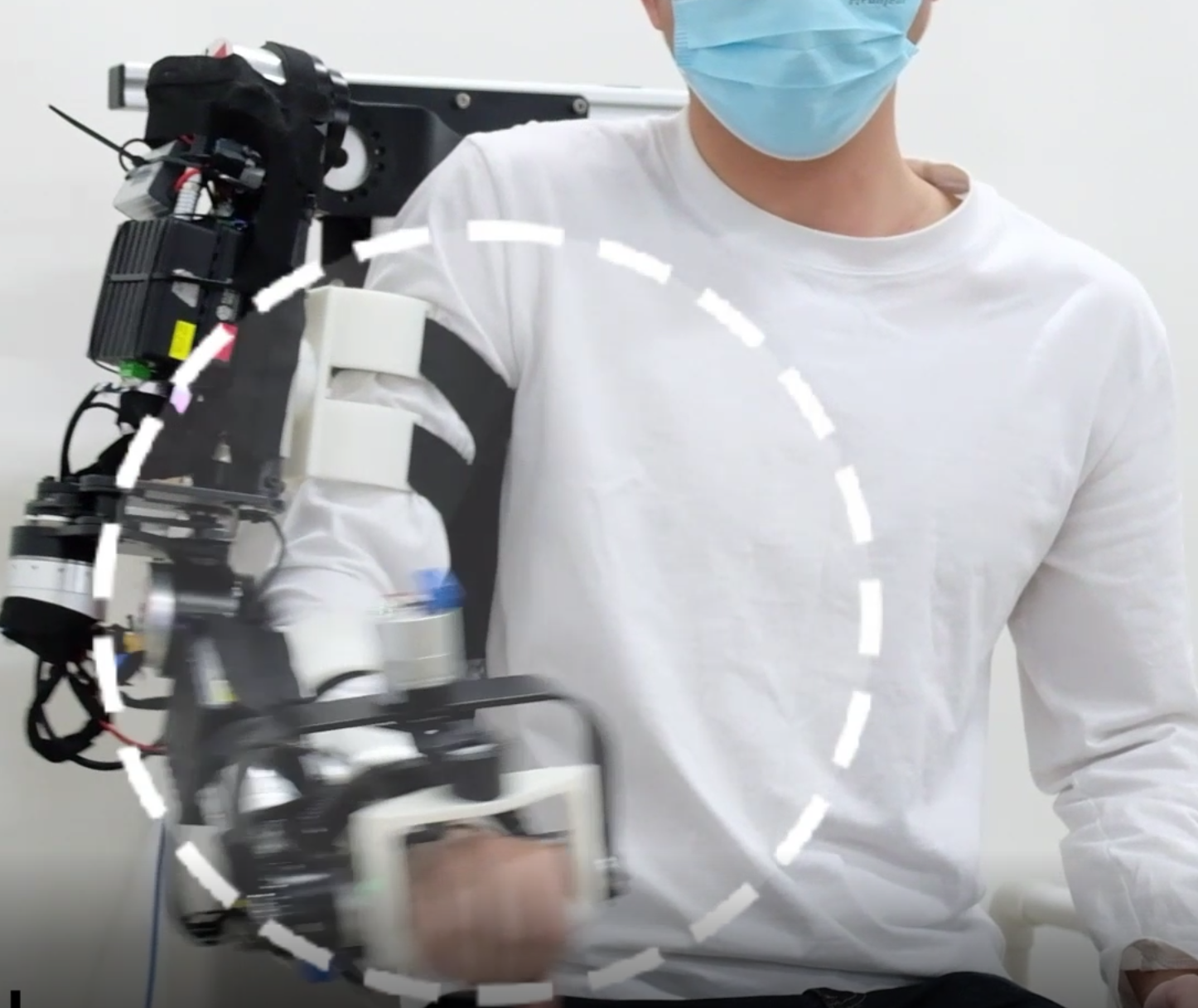}
  \caption{The operator uses the 9 DoF exoskeleton, which is implemented with the exoskeleton control framework based on a binding alignment strategy and a full-arm coordination mechanism proposed in this article, to achieve flexible and coordinated control performance during high-dynamic, large-scale movements of the entire arm.}
  \label{fig:main}
  \vspace{-0.7cm}
\end{figure}
Recently, the traditional decoupling control method involves feedforward compensation with feedback control based on interactive force~\cite{Fabian2018Automatisierungstechnik}. Regarding the feedback control, many methods are directly based on statics, calculating torque control for exoskeleton motors by multiplying the force and torque~(F/T) sensor by the transpose of the Jacobian matrix of the current binding point~\cite{zimmermann2019ral}. With this traditional method, exoskeletons can achieve general control performance, so research focuses mainly on controlling exoskeletons at the application layer rather than on the basic flexible control for exoskeletons. e.g., in rehabilitation robots~\cite{Zeiaee2022ral, Zarrin2024ral, Pan2022ral, Pan2023tnsre}, researchers focus on how to adjust assistance and resistance more conveniently to help patients better recover during training; in teleoperation applications, they focus on achieving one-sided teleoperation or bilateral force feedback teleoperation~\cite{Hejrati2023ral}; and in powered exoskeletons, providing assistance based on human intent will receive more attention~\cite{Missiroli2022ral, Grazi2022ral}.

However, in scenarios that involve high dynamic movements of the whole arm, for example, when a person flexibly operates a remote robot through an exoskeleton~\cite{rebelo2014ram} or when the exoskeleton serves as a motion data collection tool for rapid and flexible human movements, the traditional decoupling control method mentioned above~\cite{Fabian2018Automatisierungstechnik} may not work well due to uncertainties in the attachment device. This can result in jerky movements and even pose safety risks. However, few studies propose solutions from the perspective of control methods. Therefore, there is still considerable opportunity to enhance the basic flexible control performance of exoskeletons.

Regarding flexible exoskeleton control, the research team from \textit{ETH Zürich} has conducted a series of works on exoskeletons. In 2017, they compared the pros and cons of feedforward control versus disturbance observation in their designed exoskeleton AMRin\cite{Fabian2018Automatisierungstechnik}.
In 2021, they introduced a control method for the 6-DoF shoulder elbow exoskeleton AnyExo~\cite{zimmermann2020iros} that combines the acceleration information of the IMU with inverse dynamics, achieving high transparency and fast exoskeleton motion control~\cite{zimmermann2019ral}. 
In 2023, this control method was also implemented in the full-arm exoskeleton AnyExo2.0\cite{zimmermann2023tro}, achieving smooth and transparent control.

The binding attachment between the arm and the exoskeleton is critical to the exoskeleton's wearability and dynamic control performance, as sensors are usually located there. However, the comfort of wearing is frequently inversely related to the stability of the donning, which makes it difficult to achieve absolute alignment. 
In 2023, Zimmermann~\etal
~\cite{Zimmermann2023tro2} proposed a new binding device that can adaptively align different thicknesses of the arms for wearing, providing an outstanding hardware reference design for this issue. However, they tend to address problems more from the mechanical perspective, slightly increasing costs and complexity. Therefore, this has motivated us to focus on a control perspective in decreasing donning disturbance and enhancing the flexible movement performance of the exoskeleton, which can adapt to most exoskeletons equipped with the common binding attachment. There have not been any attempts in this area yet.

This paper provides a binding alignment strategy~(BAS) and a full-arm coordination mechanism~(FCM) to decrease the donning disturbance caused by binding deviations and enhance the coordination of the exoskeleton following the user's movements of the entire arm. The experimental results show that the control framework formed by both methods helps improve various exoskeleton performances.
% ~(including flexibility, adaptability, speed, accuracy, and fatigue). 

The main contributions of this paper are threefold.
\begin{itemize}
    \item We propose a comprehensive exoskeleton control framework. Based on classifying the interactive force into the major, assistant and coordination components, the control method integrates the binding alignment strategy~(BAS)~and the full-arm coordination mechanism~(FCM)~to reduce donning disturbances and improve coordination and flexibility. In addition, an intention distinction module is introduced to resolve conflicts in force sensor measurements and classify different motion intentions.
    \item We analyze potential disturbances, including the force sensor measurement deviation model due to a single wear point offset, the coupling effects of the donning offset of two binding points, and the unstable phenomena during the whole arm movement, which provides the basis for the design of the control framework. 
    \item  The method motivates us to develop a full-arm exoskeleton that covers all 9 degrees of freedom (DoF) of the human upper limb for implementation. We conduct a series of experiments to validate the advantages of our control method regarding the flexibility, adaptability, speed, accuracy, and fatigue of the exoskeleton. 
\end{itemize}

%%%%%%%%%%%%%%%%%%%%%%%%%%%%%%%%%%%%%%%%%%%%%%%%%%%%%%%%%%%%%%%%%%%%%%%%%%%%%%%%
\vspace{-0.3cm}
\section{System Description}
\vspace{-0.2cm}
\label{sec:related}

\begin{figure}[!t]
  \centering
  \includegraphics[width=1\linewidth]{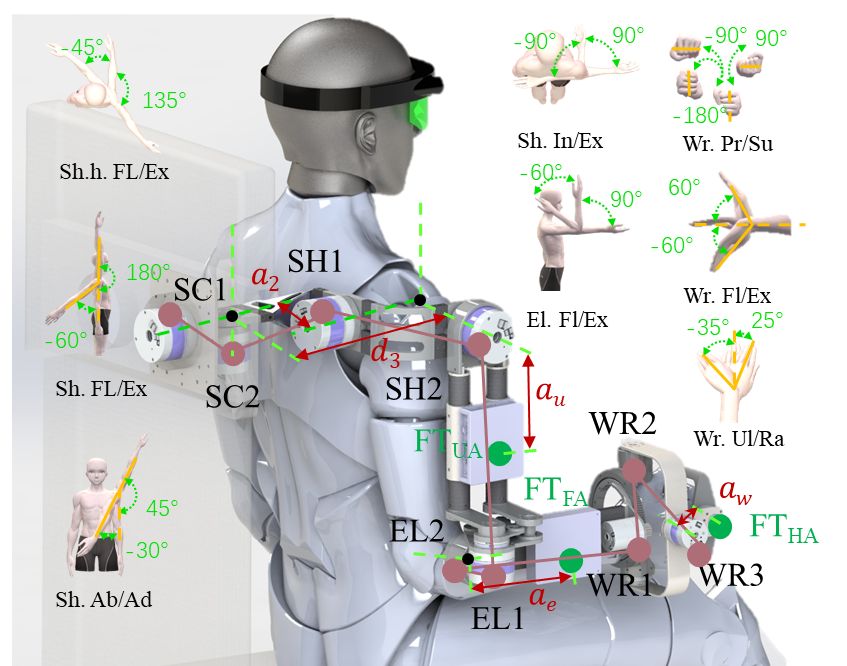}
  \caption{Overview of the exoskeleton system. The exoskeleton is driven by nine motors, which include shoulder joint motors SC1, SC2, SH1, SH2; elbow joint motors EL1, EL2; and wrist joint motors WR1, WR2, WR3. $\text{FT}_\text{UA}$, $\text{FT}_\text{FA}$, and $\text{FT}_\text{HA}$ are F/T sensors at the upper arm, forearm, and hand, respectively.}
  \label{fig:system}
  \vspace{-0.5cm}
\end{figure}

Our proposed control framework is implemented on a 9-DoF exoskeleton designed for teleoperation named NuBot\_Rafiki~\cite{Cheng2024tmech}, as shown in~\figref{fig:system}.  
It consists of the shoulder, elbow, and wrist parts. The shoulder part consists of scapular motors SC1 and SC2 that compensate for the movement of the sternoclavicular joint, as well as corresponding shoulder joint motors SH1 and SH2, which enable shoulder flexion/extension~(Sh.Fl/Ex), adduction/abduction~(Sh.Ad/Ab) and horizontal flexion/extension ~(Sh.h.Fl/Ex) movements. The motors of the elbow part EL1 and EL2 correspond shoulder interior/exterior rotation~(Sh.Ir/Er) and elbow flexion/extension~(El.Fl/Ex). The wrist part comprises three motors, WR1, WR2, and WR3, in an orthogonal structure, accompanying the wrist pronation/supination~(Wr.Pr/Su), flexion/extension~(Wr.Fl/Ex), and ulnar/radial deviation~(Wr.Ud/Rd) movements. All motors are equipped with harmonic gears and feature a 2-2-2-3 layout.
Three F/T sensors of different sizes (KunWei$^\circledR$ KW57, KW46, KW36) are installed at the upper arm~(UA), forearm~(FA), and hand attachment~(HA) to measure the interaction force between the user and the exoskeleton. It is also a common configuration for the upper limb exoskeletons~\cite{zimmermann2019ral,kim2017ijrr}.
For better measurement accuracy, we choose the ranges 50\,N and 5\,Nm in the linear and rotational directions of XYZ, and the measurement error is within 0.02\,$\%$. The drive motor and the F/T sensor use a CAN bus for signal transmission, and the data acquisition and the controller control frequency are 80\,Hz.

Previously, exoskeleton control was decoupled into shoulder, elbow, and wrist segments for separate control. 
It has an acceptable effect on the individual movement of the joints in the three parts. 
% It is instrumental in applying rehabilitation robots, where specific joint injuries often require individual joint recovery. 
But now, our control algorithm aims to improve the coordination and flexibility of movements in the whole arm and reduce the impact of uncertainty at the binding points in this generalized exoskeleton layout. Next, we will analyze the force disturbance caused by donning offsets. 

\vspace{-0.05cm}
%%%%%%%%%%%%%%%%%%%%%%%%%%%%%%%%%%%%%%%%%%%%%%%%%%%%%%%%%%%%%%%%%%%%%%%%%%%%%%%%
\section{Force Interaction Analysis at the Binding Points}
\label{sec:problem}
\vspace{-0.25cm}

\begin{figure}[t]
  \centering
  \includegraphics[width=1\linewidth]{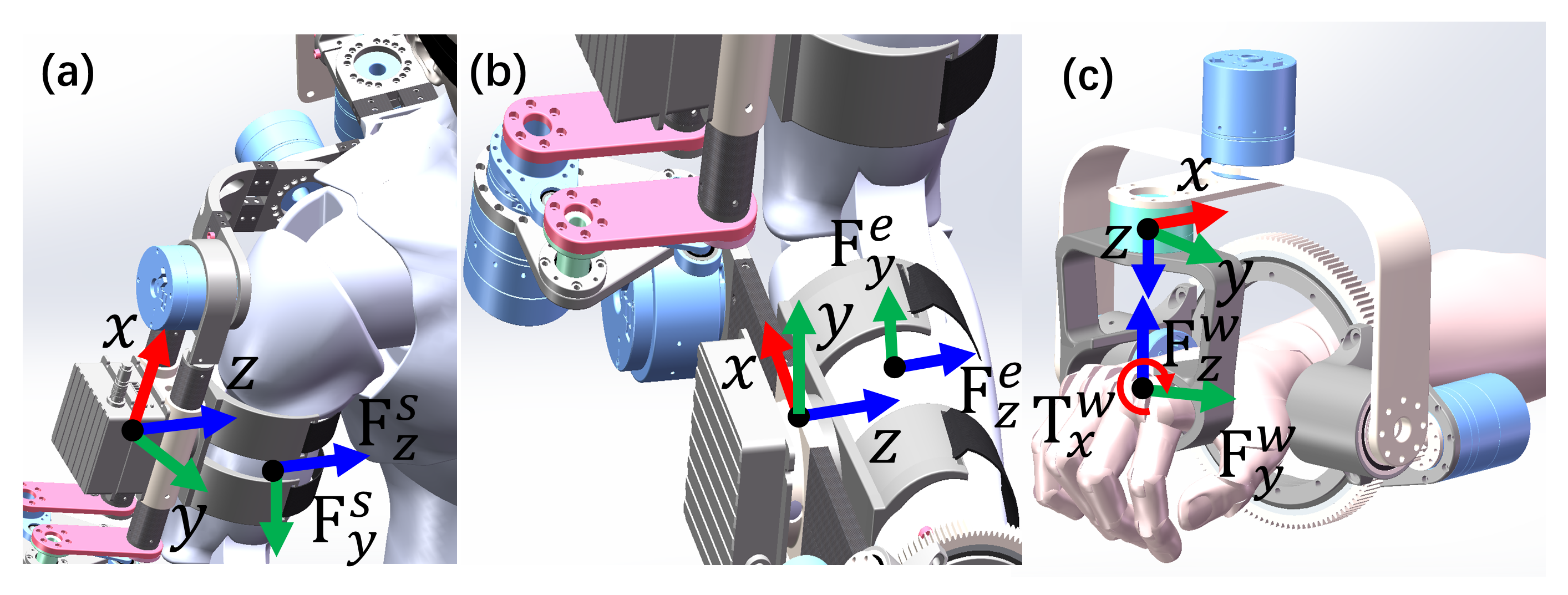}
  \caption{ In the ideal state, the interaction forces between the exoskeleton and the human at attachment points.~(a),~(b),~(c) are respectively for the upper arm, forearm, and hand binding attachment.}
  \label{fig:ideally_attach}
  \vspace{-0.5cm}
\end{figure}

Our control algorithm aims to improve the coordination and flexibility of movements in the whole arm and reduce the impact of uncertainty at the binding points. In this section, we analyze the force interaction caused by donning offsets.
The interaction force in this ideal state is illustrated in~\figref{fig:ideally_attach}. The exoskeleton fits the human body seamlessly and the attachments fit closely to the body. In the upper arm binding attachment~(UA, \figref{fig:ideally_attach}~(a)), the two directions interaction force~$F^{s}_{y}$ and $F^{s}_{z}$ corresponds to the movement of the shoulder part. In the forearm attachment~(FA, \figref{fig:ideally_attach}~(b)), the two directions force~$F^{e}_{y}$ and $F^{e}_{z}$ correspond to the movement of the elbow. In the hand attachment~(HA, \figref{fig:ideally_attach}~(c)), there are the three directions~$F^{w}_{y}$,\,$F^{w}_{z}$ and $T^{w}_{x}$ corresponding to the three joints of the wrist. 
However, in practice, there are several issues affecting the measurement of interactive forces: (1) The non-ideal installation of the F/T sensors, which cannot be positioned exactly at the point where the interactive force is generated; (2) The exoskeleton and the binding device do not fully adapt to the human body; and (3) During high-speed movements, deformation and displacement occur at the binding connection points. These factors result in deviations in the sensors' measurements of interactive forces, hindering the accurate detection of ideal forces.
To better understand the disturbances at the attachment points, we analyzed the interaction force from several aspects below.

The force interaction analysis at a single binding point is relatively straightforward. The contact force between human arms and the attachments typically acts at the edge of the attachment rather than at its center. Therefore, it will result in redundant forces and torques. The more comprehensive analysis refers to the supplementary material~(A).
\begin{figure}[t]
  \centering
  \includegraphics[width=0.9\linewidth]{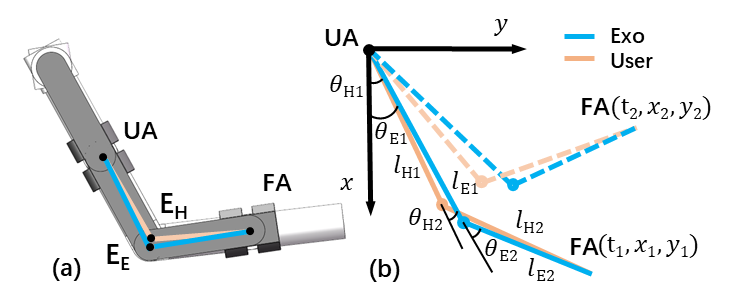}
  \caption{(a) is the simplified model diagram illustrating mismatched sizes for wearing the exoskeleton on the user's upper arm and forearm. It is similar to two linked rods moving together in a sleeve. 
  The model is further simplified to the rod model in (b). When the upper arm and forearm move together, there is a misalignment at the binding point between the upper arm and forearm rods of the user and the exoskeleton.}
  \label{fig:couple_disturbance}
  \vspace{-0.5cm}
\end{figure}
\vspace{-0.5cm}
\subsection{The coupling effects of the donning offset}
In addition to the donning offset at a single point, we observed coupling effects at different binding point offsets. Specifically, these effects primarily involve the influence of the UA binding point offset on the interactive force experienced by FA, as well as the effect of the FA binding point offset on the interactive force exerted on UA.
As shown in \figref{fig:couple_disturbance}(a), we simplify the user arm as two connected circular rods and the exoskeleton as cylindrical rods formed at the binding points. 
$\text{E}_\text{H}$ and $\text{E}_\text{E}$ refer to the rotation centers of the user's elbow and the exoskeleton's elbow, respectively. 
In practical wear situations, $\text{E}_{\text{H}}$ and $\text{E}_{\text{E}}$ may deviate due to differences in arm length. 
\begin{figure*}[tbp]
\vspace{-0.2cm}
    \begin{minipage}{0.42\linewidth}
    \captionof{table}{Causes of disturbance in unstable phenomenon}
    \footnotesize
    \begin{tabular}{p{0.2cm}|p{1cm}|p{5cm}}
    \hline
    \addlinespace[3pt]
        \multirow{5}*{\!\!\!\!UA}  &  \multirow{2}*{\begin{tabular}[c]{@{}c@{}}Flexion\\ \figref{fig:disturbance_analysis}~(b)\end{tabular}} &  $F_{y+}^{s}$ $\uparrow$, the exoskeleton $\nwarrow$, $F_{y-}^{s}$~$\uparrow$, $T_{z+}^{s}$~$\uparrow$, the exoskeleton~$\searrow$, causing vibration.  \\
        \addlinespace[3pt]
        \cline{2-3}
        \addlinespace[3pt]
          ~ & \multirow{3}*{\begin{tabular}[c]{@{}c@{}}Extension\\ \figref{fig:disturbance_analysis}~(c)\end{tabular}} & $F_{y-}^{s}$ $\uparrow$, the exoskeleton $\searrow$, $F_{y+}^{s}$ $\uparrow$, $T_{z-}^{s}$ $\uparrow$, the exoskeleton $\nwarrow$,, causing vibration.\\
          \addlinespace[3pt]
        \hline
        \addlinespace[3pt]
        \multirow{5}*{\!\!\!\!FA}  &  \multirow{3}*{\begin{tabular}[c]{@{}c@{}}Flexion\\ \figref{fig:disturbance_analysis}~(c)\end{tabular}} & $F_{y+}^{e}$ $\uparrow$, the exoskeleton $\nwarrow$, $F_{y-}^{e}$ $\uparrow$, $T_{z+}^{e}$ $\uparrow$, the exoskeleton $\searrow$, causing vibration.  \\
        \addlinespace[3pt]
        \cline{2-3}
        \addlinespace[3pt]
          ~ & \multirow{2}*{\begin{tabular}[c]{@{}c@{}}Extension\\ \figref{fig:disturbance_analysis}~(b)\end{tabular}} & $F_{y-}^{e}$ $\uparrow$, the exoskeleton $\searrow$, $F_{y+}^{e}$ $\uparrow$, $T_{z-}^{e}$ $\uparrow$, the exoskeleton $\nwarrow$, causing vibration.\\
          \addlinespace[3pt]
        \hline
    \end{tabular}
    \footnotesize{$\uparrow$~indicates a numerical increase.~$\downarrow$~indicates a numerical decrease.~$\searrow$~and~$\nwarrow$~represent the downward swing and the upward swing of the exoskeleton.}
    \label{tab:problem}
    \end{minipage}
    \begin{minipage}{0.56\linewidth}
    \vspace{-0.3cm}
    \hspace{0.3cm}\!\!\!\!\includegraphics[width=1\linewidth]{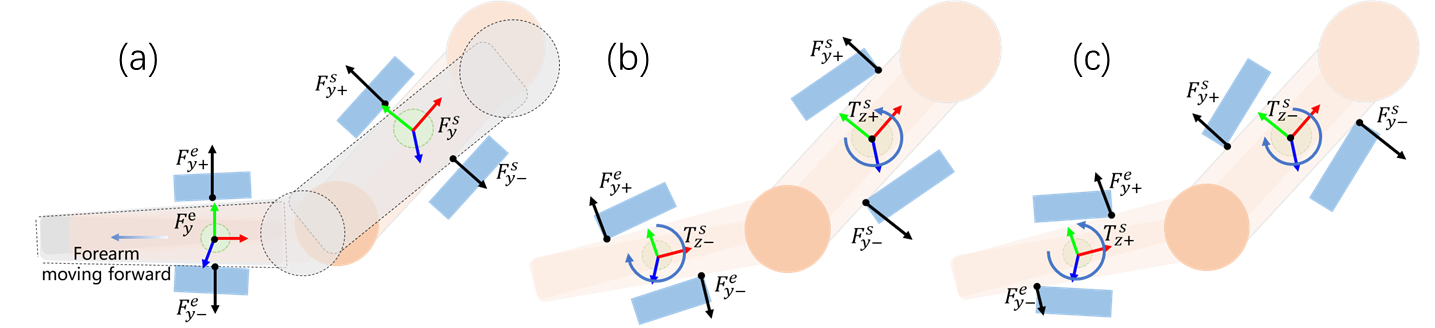}
    \vspace{0.3cm}
    \caption{Binding deviation is easily caused during whole arm movement. (a) An example of whole arm movement is from the perspective of human goals. (b) and (c) represent two different disturbance situations, respectively.}
    \label{fig:disturbance_analysis}
    \end{minipage}
    \vspace{-0.5cm}
\end{figure*}
Furthermore, we simplify this problem by considering the common motion of two connected rods with slightly different lengths, as shown in \figref{fig:couple_disturbance}(b).
From kinematics, we can obtain the relationship between the angle deviation of UA and FA, the movement position of the arm, and the arm length of the user and the exoskeleton.
$\theta_{\text{H1}}$ and $\theta_{\text{E1}}$ represent the angles between the upper arm of the user and the upper arm of the exoskeleton with the vertical plane. $\theta_{\text{H2}}$ and $\theta_{\text{E2}}$ represent the elbow angles of the human and the exoskeleton. These can be calculated using the cosine theorem; refer to the supplementary material~(B) for the specific procedure and results.
The relationship between forearm velocity and joint rotation speed can be determined from kinematics, 
\begin{equation}
\left[\begin{array}{l}
\mathrm{d} x \\
\mathrm{d} y
\end{array}\right]=\boldsymbol{J}\boldsymbol{\theta}=\left[\begin{array}{cc}
-l_1 {\rm S}1 -l_2 {\rm S}12 & -l_2 {\rm S}12 \\
l_1 {\rm C}1 + l_2 {\rm C}12 & l_2 {\rm C}12
\end{array}\right]\left[\begin{array}{l}
\mathrm{d} \theta_1 \\
\mathrm{d} \theta_2
\end{array}\right],
\label{equ:J-1}
\end{equation}
where~${\rm S}1$, ${\rm C}1$ denote~$\sin(\theta_1)$~and~$\cos(\theta_1)$,~${\rm S}12$~and~${\rm C}12$ denote~$\sin(\theta_1+\theta_2)$~and~$\cos(\theta_1+\theta_2)$ respectively. The inverse of the Jacobian matrix is
\begin{equation}
\left[\begin{array}{cc}
{\rm C}12/(l_1 {\rm S}2) & {\rm S}12/(l_1 {\rm S}2) \\
-(l_2 {\rm C}12 + l_1 {\rm C}1)/(l_1 l_2 {\rm S}2) & -(l_2 {\rm S}12 + l_1 {\rm S}1)/(l_1 l_2 {\rm S}2)
\end{array}\right].
\label{equ:J-2}
\end{equation}
%因此在绑缚点产生的扭转角偏差与该偏差的速度都可以求出来
Then, by substitute $\theta_{\text{H1}}$, $\theta_{\text{H2}}$ and $\theta_{\text{E1}}$, $\theta_{\text{E2}}$ into the above equations respectively, the torsional angle deviation $\theta_{\text{\tiny UA}}^{err}$ and $\theta_{\text{\tiny FA}}^{err}$ generated at the binding point and its speed $\dot{\theta}_{\text{\tiny UA}}^{err}$ and $\dot{\theta}_{\text{\tiny FA}}^{err}$ can be calculated as
\begin{equation}
\left\{\begin{array}{l}
\theta_{\text{\tiny UA}}^{err}=\theta_{\text{E1}} - \theta_{\text{H1}}\\
\theta_{\text{\tiny FA}}^{err}=\theta_{\text{E1}} + \theta_{\text{E2}} - \theta_{\text{H1}} - \theta_{\text{H2}}\\
\end{array}\right.
\end{equation}
\begin{equation}
\left[\begin{array}{c}
\dot{\theta}_{\text{\tiny {UA}}}^{err} \\
\dot{\theta}_\text{\tiny {FA}}^{err}
\end{array}\right]=\left(\boldsymbol{J}_\text{\tiny {UF}}^{\mathrm{E}^{-1}}-\boldsymbol{J}_\text{{\tiny UF}}^{H^{-1}}\right)\left[\begin{array}{l}
x \\
y
\end{array}\right]+\left[\begin{array}{c}
0 \\
\dot{\theta}_{\text{\tiny UA}}^{err}
\end{array}\right]
\end{equation}
${\boldsymbol{J}_\text{\tiny{UF}}^{E^{-1}}}$~and~${\boldsymbol{J}_\text{\tiny {UF}}^{H^{-1}}}$ denote the first row of the inverse matrix of the Jacobian matrix from binding point UA to binding point FA of the exoskeleton and the human, respectively.  They can be obtained from~\eqref{equ:J-1}~and~\eqref{equ:J-2}.
The disturbance torque generated is represented using the impedance model\cite{Zimmermann2023tro2}:
\vspace{-0.2cm}
\begin{equation}
\begin{array}{l}
T_{\text{\tiny UA}}^{err}= K_{\text{\tiny UA}} \theta_{\text{\tiny UA}}^{err} + D_{\text{\tiny UA}} \dot{\theta}_{\text{\tiny UA}}^{err},\\ 
T_{\text{\tiny FA}}^{err}= K_{\text{\tiny FA}} \theta_{\text{\tiny FA}}^{err} + D_{\text{\tiny FA}} \dot{\theta}_{\text{\tiny FA}}^{err}\\
\end{array}
\end{equation}
$K_{\text{\tiny UA}}$ and $K_{\text{\tiny FA}}$ are the stiffness coefficients at binding points UA and FA, respectively. $D_{\text{\tiny UA}}$ and $D_{\text{\tiny FA}}$ are the damping coefficients of the two binding points\cite{Zimmermann2023tro2}. 
Therefore, it is necessary to consider reducing the impact of~$T_{\text{\tiny UA}}^{err}$~and~$T_{\text{\tiny FA}}^{err}$~in the subsequent control method design.
\vspace{-0.3cm}
\subsection{Qualitative analysis of another unstable phenomenon.}
\vspace{-0.2cm}
In addition to the analysis above, we found that during the traditional decoupling control method, force disturbances can lead to oscillations or vibrations occurring in extreme states.
In the traditional decoupling control method, the F/T sensor data is used to directly control the respective joints of the shoulder, forearm, and hand. However, when humans control their arms, there are two perspectives: joint-oriented motion and target-oriented motion.
For example, when a person intends to extend their forearm forward, he considers directly extending the forearm forward rather than adjusting the elbow and shoulder positions. \figref{fig:disturbance_analysis}~(a) shows the force interactions with the forearm intentionally extending forward. 
And because the users will not consider their own decoupled motion of the elbow and shoulder during target-oriented movements, such as forearm forward extending, the traditional decoupling control approach tends to lead to binding offsets and misalignments. e.g., \figref{fig:disturbance_analysis}~(b)~and~(c) show the two unstable situations. Different upper arm and forearm movements will further lead to extremely unstable movement situations in both cases, as qualitatively analyzed in \tabref{tab:problem}.

In the scenario depicted in~\figref{fig:disturbance_analysis}~(b), if the whole arm is extended forward, there is an upward swing and a downward swing intention of the upper arm and the forearm. Regarding the upper arm upward~(Flexion), 
$F_{y+}^{s}$ increases, causing the exoskeleton to swing upward, then $F_{y-}^{s}$ increases, leading to an increase in $T_{z+}^{s}$, which results in a downward swing~(Extension) and subsequent vibrations. 
Regarding the forearm downward~(Extension) at this moment, 
$F_{y-}^{e}$ starts to increase, causing the exoskeleton to swing downward, then $F_{y+}^{s}$ increases, leading to an increase in $T_{z-}^{e}$, which results in an upward swing~(Flexion) and subsequent vibrations. 

In the scenario depicted in~\figref{fig:disturbance_analysis}~(c), if the whole arm is retracted backward, the upper arm swings upward, while the forearm swings downward.
Regarding the upper arm downward (Extension), 
$F_{y-}^{s}$ increases, followed by an increase in $F_{y+}^{s}$ which may exceed the former. Although $T_{z-}^{s}$ also increases, this can cause the exoskeleton to swing downward, creating resistance and oscillating upward and downward movements.
Regarding the forearm upward~(Flexion), 
$F_{y-}^{e}$ increases, causing the exoskeleton to swing upward. As $F_{y+}^{e}$ and $T_{z+}^{s}$ rise, the exoskeleton swings downward, resulting in vibrations.

Therefore, there is a critical need for a method that not only resolves the coordination between joint-oriented intention movements and target-oriented movements in exoskeletons but also mitigates motion disturbances caused by donning deviations at the binding points. 
\vspace{-0.3cm}
\subsection{Sensor data classification}
\vspace{-0.1cm}
Through the analysis in the above subsections, we recognize the impact of different force components from the three F/T sensors on exoskeleton movement. Therefore, we categorize the measurements from the three binding points' force sensors into different types, as shown in \tabref{tab:force_component}. 
First, the measured interactive force data are divided into four categories: 
\begin{table}[t]
    \centering
    \captionof{table}{The classification of exoskeleton force sensor data.}
    \vspace{-0.2cm}
    \resizebox{9cm}{!}{
    \begin{tabular}{cc|c|c|c}
    \toprule[1.5pt]
    \multicolumn{2}{c|}{Binding point} & UA & FA & HA \\
    \midrule[1pt]
        \multicolumn{2}{c|}{\footnotesize{Major component~(MC)}} & $F^{s}_{y}$, $F^{s}_{z}$ & $F^{e}_{y}$, $F^{e}_{z}$ & $T^{w}_{x}$, $F^{w}_{y}$, $F^{w}_{z}$\\
        \multicolumn{2}{c|}{\footnotesize{Assistant component~(AC)}} & $T^{s}_{y}$, $T^{s}_{z}$ & $T^{e}_{y}$, $T^{e}_{z}$ & $T^{w}_{y}$, $T^{w}_{z}$\\ 
        \cline{0-1}
        \footnotesize{Coordination} & \footnotesize{uncouple} & $F^{s}_{x}$ & $F^{e}_{x}$ & $F^{w}_{x}$\\
        \cline{2-2}
         \footnotesize{component~(CC)}  & \footnotesize{couple} &   & $F^{e}_{y}$, $F^{e}_{z}$ & $F^{w}_{y}$, $F^{w}_{z}$\\
        \cline{0-1}
        \multicolumn{2}{c|}{\footnotesize{Redundant component~(RC)}}  & $T^{s}_{x}$ & $T^{e}_{x}$ &    \\
    \bottomrule[1.5pt]
    \end{tabular}
         }
    \label{tab:force_component}
    \vspace{-0.5cm}
\end{table}
\begin{figure*}[t]
\vspace{-0.5cm}
  \centering
   \includegraphics[width=0.81\linewidth]{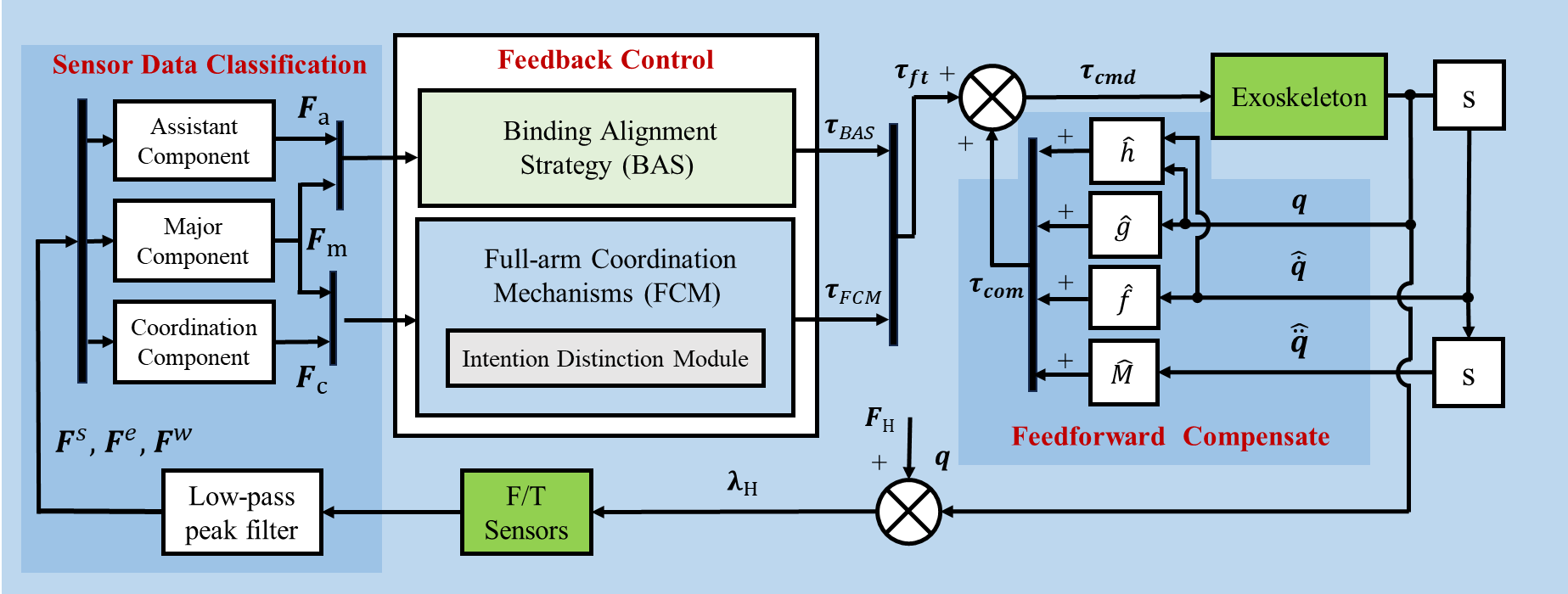}
    \captionof{figure}{Control diagram of the exoskeleton system. $\boldsymbol{F}_{\text{H}}$ denotes the human arm active force. The three F/T sensors~(at UA, FA, and HA) measure the interaction force $\boldsymbol{\lambda}_{\text{H}}$ after the low-pass filter represented by $\boldsymbol{F}^s$, $\boldsymbol{F}^e$, and $\boldsymbol{F}^w$ respectively.
    After the sensor data classification~(see~\tabref{tab:force_component}), 
    the major components~(MC) and assistant components~(AC) extracted from the interactive force data generate the alignment control variable~$\boldsymbol{\tau}\!_{_{B\!A\!S}}$ through the binding alignment strategy~(BAS). The major components and coordination components~(CC) will be input into the full-arm coordination mechanism~(FCM) to produce the coordination control variable~$\boldsymbol{\tau}\!_{_{F\!C\!M}}$. The coupled coordinated components will be processed by the intention distinction module to resolve the conflict with the major components.
    Then~$\boldsymbol{\tau}_{ft}$ is the sum of~$\boldsymbol{\tau}\!_{_{B\!A\!S}}$ and~$\boldsymbol{\tau}\!_{_{F\!C\!M}}$, which, along with the dynamics compensation ~$\boldsymbol{\tau}_{com}$~(consist of inertia~$\hat{\boldsymbol{M}}$, Coriolis force~$\hat{\boldsymbol{h}}$, gravity~$\hat{\boldsymbol{g}}$, and friction compensation~$\hat{\boldsymbol{f}}$), constitutes the control variable of the exoskeleton motors~$\boldsymbol{\tau}_{cmd}$. Encoder measures the joint position $\hat{\boldsymbol{q}}$. Then, through differentiation, the velocity $\hat{\dot{\boldsymbol{q}}}$ and acceleration $\hat{\ddot{\boldsymbol{q}}}$ are obtained.}
    \label{fig:exo_control}
    \vspace{-0.5cm}
\end{figure*}
\begin{figure}[t]
    \centering
    \includegraphics[width=0.81\linewidth]{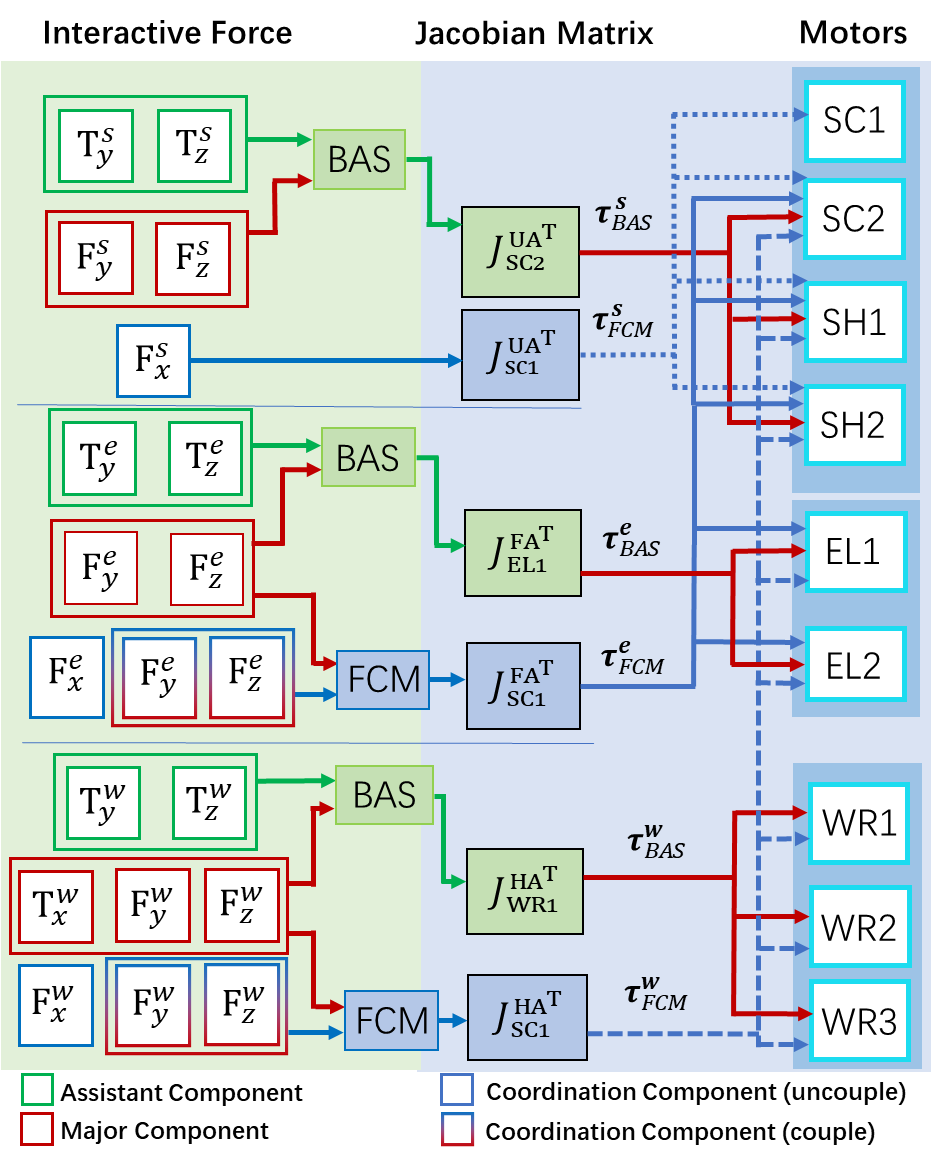}
    \caption{Detailed block diagram of the feedback control section: 
    The green area represents different sensor components, processed according to classification through BAS (see \textbf{Strategy.\,\ref{alg:ft}}) and FCM (see \textbf{Strategy.\,\ref{alg:ft2}}), multiplied by the corresponding Jacobian matrix to obtain the alignment control variables~$\boldsymbol{\tau}\!_{_{B\!A\!S}}^{{\,}s}$,~$\boldsymbol{\tau}\!_{_{B\!A\!S}}^{{\,}e}$,~$\boldsymbol{\tau}\!_{_{B\!A\!S}}^{{\,}w}$ and coordination control variables ~$\boldsymbol{\tau}\!_{_{F\!C\!M}}^{{\,}s}$,~$\boldsymbol{\tau}\!_{_{F\!C\!M}}^{{\,}e}$,~$\boldsymbol{\tau}\!_{_{F\!C\!M}}^{{\,}w}$ generated for the shoulder, elbow, and wrist motors~(see \figref{fig:system}). These will collectively influence the torque control of each motor, and the arrows in the diagram indicate their relationships.} 
    \label{fig:control}
    \vspace{-0.5cm}
\end{figure}

\textbf{Major components} are the most important collecting data, composed of force perception data generated under ideal wearing conditions, as shown in \figref{fig:ideally_attach}. 
\textbf{Assistant components} are generated by motion deviation under the user's subjective intentions. They can be used to adjust the degree of alignment of the exoskeleton with the upper limb to avoid discomfort caused by the misalignment of wearing the exoskeleton during high-speed movement. 
e.g., when torque signals $T_{z}^{e}$ and $T_{y}^{e}$ have numerical values, it indicates that the donning state is not in a perfect fit. Therefore, it is necessary to decrease the deviation at the binding points based on the assistant components.
\textbf{Coordination components} are the measured force components that can be used in our full-arm coordination mechanism from a target-oriented operation perspective. 
Some components in this category coincide with the major components at the current donning point, which we refer to as coupling components, such as $F^{e}_{y}$, $F^{e}_{z}$, $F^{w}_{y}$, $F^{w}_{z}$. Those that do not coincide with the major components are referred to as uncoupling components, such as $F^{e}_{x}$ and $F^{w}_{x}$. 
Uncoupling components can be used directly in the controller design. However, coupling components must consider the conflicts with the major components, which requires the intention distinction module we proposed.
\textbf{Redundant components} generally do not arise from active intentions, and it is difficult to utilize them for beneficial control influences. 
They are directly attenuated in the control.

After the above classification, the components 
are hierarchically implemented to construct the motor control variables diagram as shown in \figref{fig:control}, which will be described in the next section.

%%%%%%%%%%%%%%%%%%%%%%%%%%%%%%%%%%%%%%%%%%%%%%%%%%%%%%%%%%%%%%%%%%%%%%%%%%%%%%%%

\section{Control Approach}

The overall control framework for the exoskeleton consists of three parts: sensor data classification, feedforward compensation, and feedback regulation based on the active interaction forces. 
As shown in \figref{fig:exo_control}, the F/T sensors at the upper arm, forearm, and hand binding position measure the interaction forces between the human arm and the exoskeleton. 
The force and torque data for each binding point are represented by $\boldsymbol{F}^s$, $\boldsymbol{F}^e$, and $\boldsymbol{F}^w$. 
Through the sensor data classification, as shown in \tabref{tab:force_component}, the major components~(MC) $\boldsymbol{F}_m$~($\boldsymbol{F}_m^s$, $\boldsymbol{F}_m^e$, $\boldsymbol{F}_m^w$) and assistant components~(AC) $\boldsymbol{F}_a$~($\boldsymbol{F}_a^s$, $\boldsymbol{F}_a^e$, $\boldsymbol{F}_a^w$) are extracted from the interactive F/T data. They generate the alignment control variable~$\boldsymbol{\tau}\!_{_{B\!A\!S}}$ through the binding alignment strategy~(BAS).
Then, the major components and uncoupled coordination components~(CC) will be input into the full-arm coordination mechanism~(FCM) to produce the coordination control variable~$\boldsymbol{\tau}\!_{_{F\!C\!M}}$. The intention distinction module of FCM will process the coupled coordination components to resolve the conflict with the major components.
The alignment control variable~$\boldsymbol{\tau}\!_{_{B\!A\!S}}$ and the coordination control variable~$\boldsymbol{\tau}\!_{_{F\!C\!M}}$ form the feedback regulation control variable~$\boldsymbol{\tau}_{ft}$ based on the measured interactive force. 
Then, the control variables for the exoskeleton motors, denoted as~$\boldsymbol{\tau}_{cmd}$, are formed by adding the dynamics compensation~$\boldsymbol{\tau}_{com}$ to~$\boldsymbol{\tau}_{ft}$.
The dynamic compensation consists of inertia~$\hat{\boldsymbol{M}}$, Coriolis force~$\hat{\boldsymbol{h}}$, gravity~$\hat{\boldsymbol{g}}$, and friction compensation~$\hat{\boldsymbol{f}}$.
The fundamental control law can be expressed as follows:

\vspace{-0.3cm}
\begin{equation}
\boldsymbol{\tau}_{c m d} = \boldsymbol{\tau}_{com} +\boldsymbol{\tau}_{f t},
\label{equ:control_law1}
\end{equation}
\vspace{-0.3cm}
\begin{equation}
\boldsymbol{\tau}_{com} = \hat{\boldsymbol{M}}\left(\boldsymbol{q}\right) \hat{\ddot{\boldsymbol{\boldsymbol {q}}}} + \hat{\boldsymbol{h}}\left(\boldsymbol{q}, \hat{\dot{\boldsymbol{q}}}\right) +
\hat{\boldsymbol{g}}\left(\boldsymbol{q}\right) + \hat{\boldsymbol{f}}\left(\dot{\boldsymbol{q}}\right),
\label{equ:dynamic4}
\end{equation}
\vspace{-0.3cm}
\begin{equation}
\boldsymbol{\tau}_{ft} = \boldsymbol{\tau}\!_{_{B\!A\!S}} + \boldsymbol{\tau}\!_{_{F\!C\!M}}.
\label{equ:control_law3}
\end{equation}

Then, we will introduce the binding alignment strategy~(BAS) and the full-arm coordination mechanism~(FCM) with the intention distinction module, which is our key work. Finally, we will introduce the dynamic compensation of the control system.

\label{sec:Control}
\floatname{algorithm}{Strategy}
\begin{algorithm}[t]
    \caption{Binding alignment strategy~(BAS)}
    \label{alg:ft}    
    \renewcommand{\algorithmicrequire}{\textbf{Input:}}
    \renewcommand{\algorithmicensure}{\textbf{Output:}}
    \begin{algorithmic}[1]
        \Require {$\left\{\begin{array}{lllllll}
        \text{MC}: &F^{s}_{y}, &F^{s}_{z}, &F^{e}_{y}, &F^{e}_{z}, &F^{w}_{y}, &F^{w}_{z}\\
        \text{AC}: &T^{s}_{y}, &T^{s}_{z}, &T^{e}_{y}, &T^{e}_{z}, &T^{w}_{y}, &T^{w}_{z}
        \end{array}\right.$}

        \Ensure Processed components $F_{y}^{'}$, $T_{z}^{'}$

        \State  The exoskeleton tracks the motion of the human upper limb.
            \If {$|F_{y}| > |F_{y_{max}}|$}
                
            \State $F_{y}=F_{y_{max}}\text{sgn}(F_{y})$. sgn is the sign function.
            % \ELSE
            %     \STATE DDDDD
            \EndIf

            \State$k_{t}=\frac{\cos{\left|\pi\frac{F_{y}}{F_{y_{max}}}\right|}+1}{2}$

            \State$k_{f}=\frac{\sin{\left|\pi\frac{F_{y}}{F_{y_{max}}}\right|}+1}{2}$
            \State$F_{y}^{'}=k_{f}{F_{y}}$, $T_{z}^{'}=k_{t}{T_{z}}$

        \State\Return $F_{y}^{'}$, $T_{z}^{'}$
\end{algorithmic}
        \begin{tablenotes}
            \item \footnotesize{The AC $T_{y}^{i}$ and $T_{z}^{i}$ correspond to the MC $F_{y}^{i}$ and $F_{z}^{i}$~($T_{x}^{w}$ has no AC that corresponds to it.). $i$ denotes the different F/T sensors in each binding point. Here is an example using one set of MC and AC.}
        \end{tablenotes}
\end{algorithm}
\begin{figure}[ht]
    \centering
    \includegraphics[width=1\linewidth]{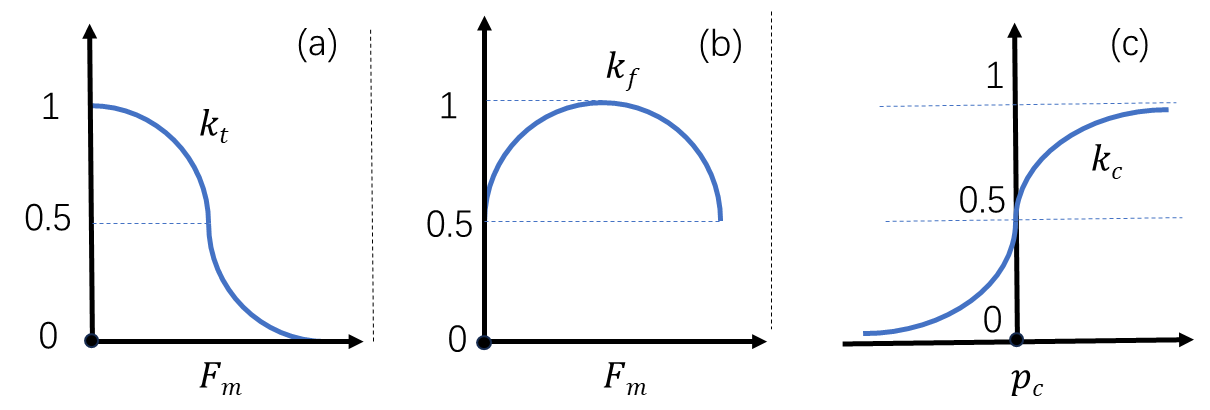}
    \caption{The curve of the nonlinear functions followed by key parameters in binding alignment strategy~(BAS) and intention distinction module~(IDM). (a), (b) are the curves showing the variation of the assistant component moment and the gain coefficient of the major component force with the numerical value of the major component. (c) shows the coordination component gain coefficient with the ratio of the numerical value of the major component at different binding points.} 
    \label{fig:curve}
    \vspace{-0.5cm}
\end{figure}
\subsection{Binding alignment strategy}
To better utilize assistant components to reduce donning deviation, we designed the binding alignment strategy~(BAS), as illustrated in \textbf{Strategy.\,\ref{alg:ft}}.
This strategy involves multiplying the force (major component) and torque (assistant component) measured at the binding point by the gain coefficients to alter the influence of the major component and assistant component on the control of the exoskeleton motors. The gain coefficients are adjusted based on the two nonlinear functions as shown in~\eqref{equ:fm}~and~\eqref{equ:tm}
with the major component~(i.e., the active intended drive force from the user) as the input variable.
The function curves are shown in~\figref{fig:curve}~(a) and (b).

First, a pair of MC and AC, such as~$F_y^s$,~$T_y^s$ measured from UA F/T sensor are input in the strategy. Next, we verify whether the MC $F_y^s$ exceeds the threshold limit~$F_{y_{max}}^s$; if it does, we set $F_y^s$ equal to $F_{y_{max}}^s$. Finally, the gain coefficients of MC and AC 
are adjusted based on the value of the major components through the two nonlinear trigonometric functions, 
\begin{equation}
    F_{y}^{'}={k_t}{F_{y}}=\frac{\sin{\left|\pi\frac{F_{y}}{F_{y_{max}}}\right|}+1}{2}{F_{y}},
    \label{equ:fm}
\end{equation}
\begin{equation}
    T_{z}^{'}={k_f}{T_{z}}=\frac{\cos{\left|\pi\frac{F_{y}}{F_{y_{max}}}\right|}+1}{2}{T_{z}}.
    \label{equ:tm}
\end{equation}
The basic idea is that when the major component of the F/T sensor at the binding point is small, the assistant component should play a larger role in quickly aligning the exoskeleton with the human arm. At the same time, the major component is attenuated to highlight the effect of the assistant component. Conversely, when the major component measured by the F/T sensor is large (i.e., when the active intention of the arm movement is significant), the torque gain coefficient~$k_t$ for the assistant component inputting to the controller gradually decreases, while the force gain for the major component gradually increases. Furthermore, when the MC is beyond the threshold, the effect of the assistant component at the binding point on the exoskeleton's movement will be diminished to zero, and the effect of the major component will also be diminished to maintain safety.
As shown in \figref{fig:control}, through the BAS process, 
\begin{equation}                    
    \boldsymbol{\tau}\!_{_{B\!A\!S}}^{{\,}s}={\boldsymbol{J}\!_{\text{\tiny{SC2}}}^{{\,}\text{\tiny{UA}}}\!\!^\top}{\boldsymbol{F}_{a}^{s}},~~
    \boldsymbol{\tau}\!_{_{B\!A\!S}}^{{\,}e}={\boldsymbol{J}\!_{\text{\tiny{EL1}}}^{{\,}\text{\tiny{FA}}}\!\!^\top}{\boldsymbol{F}_{a}^{e}},~~
    \boldsymbol{\tau}\!_{_{B\!A\!S}}^{{\,}w}={\boldsymbol{J}\!_{\text{\tiny{WR1}}}^{{\,}\text{\tiny{HA}}}\!\!^\top}{\boldsymbol{F}_{a}^{w}},
    \label{equ:tau_bas}
\end{equation}
$\boldsymbol{J}\!_{\text{\tiny{SC2}}}^{{\,}\text{\tiny{UA}}}$ represents the Jacobian matrix of the binding point UA relative to the second joint motor SC2. There are three joint motors between them, which are SC2, SH1, and SH2. The meaning of superscripts and subscripts of $\boldsymbol{J}\!_{\text{\tiny{EL1}}}^{{\,}\text{\tiny{FA}}}$ and $\boldsymbol{J}\!_{\text{\tiny{WR1}}}^{{\,}\text{\tiny{HA}}}$ follows this pattern.
They can be derived from kinematics, with specific results found in the supplementary material~(C).
Based on \textbf{Strategy.\,\ref{alg:ft}}, we have the following definitions:
\begin{equation}
\begin{array}{l}
    \boldsymbol{F}_{a}^{s}=[0, F_y^{s'}, F_z^{s'}, 0, T_y^{s'}, T_z^{s'} ]^\top,\\
    \boldsymbol{F}_{a}^{e}=[0, F_y^{e'}, F_z^{e'}, 0, T_y^{e'}, T_z^{e'} ]^\top,\\ \boldsymbol{F}_{a}^{w}=[0, F_y^{w'}, F_z^{w'}, T_x^{w}, T_y^{w'}, T_z^{w'}]^\top. 
\end{array}
\end{equation}
Therefore, we obtain the control variable generated by BAS,
\begin{equation}
\boldsymbol{\tau}\!_{_{B\!A\!S}} =[{\boldsymbol{\tau}\!_{_{B\!A\!S}}^{{\,\,}s}}^{{\!\!\!\!}\top}, {\boldsymbol{\tau}\!_{_{B\!A\!S}}^{{\,\,}e}}^{{\!\!\!\!}\top}, {\boldsymbol{\tau}\!_{_{B\!A\!S}}^{{\,\,}w}}^{{\!\!\!\!}\top}]^\top.
\label{equ:tau_a}
\end{equation}

\subsection{Full-arm coordination mechanism}
Humans typically exhibit two distinct patterns of movement intention: the joint-oriented movement and the target-oriented movement when performing upper limb actions.
For instance, during fitness exercises, they focus on specific upper limb joints to strengthen their muscles, which belongs to the joint-oriented movement. They may also focus on the objects or their hands with a clear target in mind, and the movement involves the entire arm, such as picking and placing, which belongs to the target-oriented movement.

Our full-arm coordination mechanism leverages this phenomenon. For example, when controlling exoskeletons to take the forward movement of the forearm~\figref{fig:disturbance_analysis}~(a), in the traditional decoupling method, controlling the SE1 and SE2 joints is linked only to FA F/T sensor data, while the SC1-SC2-SH1-SH2 is related solely to UA F/T sensor data. By employing the full-arm coordination mechanism, the F/T sensor data of the binding point of the hand, forearm, and upper arm are not only used to control their respective joint motors but can also be used to generate coordination control variables for movements targeting the hand or forearm as the main intention of movement target of the exoskeleton. As shown in \figref{fig:control},
\begin{equation}                    
    \boldsymbol{\tau}\!_{_{F\!C\!M}}^{{\,}s}={\boldsymbol{J}\!_{\text{\tiny{SC1}}}^{{\,}\text{\tiny{UA}}}}^{\!\top}{\boldsymbol{F}_{c}^{s}},~~
    \boldsymbol{\tau}\!_{_{F\!C\!M}}^{{\,}e}={\boldsymbol{J}\!_{\text{\tiny{SC1}}}^{{\,}\text{\tiny{FA}}}\!^\top}{\boldsymbol{F}_{c}^{e}},~~
    \boldsymbol{\tau}\!_{_{F\!C\!M}}^{{\,}w}={\boldsymbol{J}\!_{\text{\tiny{SC1}}}^{{\,}\text{\tiny{FA}}}\!^\top}{\boldsymbol{F}_{c}^{w}},
    \label{equ:tau_c_s}
\end{equation}
where~$\boldsymbol{J}\!_{\text{\tiny{SC1}}}^{{\,}\text{\tiny{UA}}}$, $\boldsymbol{J}\!_{\text{\tiny{SC1}}}^{{\,}\text{\tiny{FA}}}$, $\boldsymbol{J}\!_{\text{\tiny{SC1}}}^{{\,}\text{\tiny{FA}}}$ denote Jacobian matrix of the current binding point relative to the exoskeleton base. $\boldsymbol{F}_{c}^{s}$, $\boldsymbol{F}_{c}^{e}$, $\boldsymbol{F}_{c}^{w}$ represents the target-oriented forces at these three binding points respectively, primarily composed of the coordinating components as shown in \tabref{tab:force_component}. 
\begin{equation}
    \begin{array}{l}
    \boldsymbol{F}_{c}^{s}=[F_x^{s}, 0, 0, 0, 0, 0 ]^\top,\\
    \boldsymbol{F}_{c}^{e}=[F_x^{e}, F_y^{e}, F_z^{e}, 0, 0, 0 ]^\top,\\
    \boldsymbol{F}_{c}^{w}=[F_x^{w}, F_y^{w}, F_z^{w}, 0, 0, 0 ]^\top.
    \end{array}
\end{equation}
When determining $\boldsymbol{F}_{c}^{e}$, $\boldsymbol{F}_{c}^{w}$, we found there is a conflict between the major components and the coordinating components. This also involves whether the current user is performing joint-oriented or target-oriented intent.
Therefore, we introduce the intention distinction module~(IDM) to solve the conflict.
e.g., when the F/T sensor in FA measures components $F_x^e$, $F_y^e$, and $F_z^e$, the challenge lies in determining whether they are the major components for decoupling elbow joint movement or the coordinate components for the target-oriented intent, with the forearm as the target. The basic idea is shown in \textbf{Strategy.\,\ref{alg:ft2}}. 

\floatname{algorithm}{Strategy}
\begin{algorithm}[t]
    \caption{Intention distinction module in the full-arm coordination mechanism~(FCM) }
    \label{alg:ft2}
    \renewcommand{\algorithmicrequire}{\textbf{Input:}}
    \renewcommand{\algorithmicensure}{\textbf{Output:}}
    \begin{algorithmic}[1]
        \Require F/T data $\boldsymbol{F}^{s}$, $\boldsymbol{F}^{e}$, $\boldsymbol{F}^{w}$
        \Ensure ${\boldsymbol{F}_{c}^{e}}^{'}$    %%output
        
        \State  The exoskeleton tracks the motion of the human upper limb.
        \If {$((F_{y}^{e} < F_{y_{\_{th}}}^{e})~\textbf{and}~(F_{z}^{e} > F_{z_{\_{th}}}^{e}))~\textbf{or}~((F_{y}^{s} < F_{y_{\_{th}}}^{s})~\textbf{and}~(F_{z}^{s} < F_{z_{\_{th}}}^{s}))$} 
        %当前绑缚处以上的绑缚处传感器主分量很小,但当前绑缚处的主分量比较大
            \State $S_m\in\mathbb{J}$. Indicate the user's intention for joint movement.
            \State $\boldsymbol{\tau_{c}}^{e} =0 $ 
        \ElsIf {$((F_{y}^{e} \geq F_{y_{\_{th}}}^{e})~\textbf{or}~(F_{z}^{e} > F_{z_{\_{th}}}^{e}))~\textbf{and}~((F_{y}^{s} > F_{y_{\_{th}}}^{s})~\textbf{or}~(F_{z}^{s} \geq F_{z_{\_{th}}}^{s}))$}
            \State {$S_m\in\mathbb{T}$}. Indicate the user's intention for target-oriented movement.%处于以目标为导向的操作意图
            \State  $a_e=\frac{\|{\boldsymbol{F}_{mc}^{e}}\|_2}{\|{{\boldsymbol{F}_{mc}^{e}}_{\_c}}\|_2}, a_s=\frac{\|{\boldsymbol{F}_{mc}^{s}}\|_2}{\|{{\boldsymbol{F}_{mc}^{s}}_{\_c}}\|_2}$.
            \If {$\frac{a_e}{a_s}<1$}
                \State $p_c^e=-a_s/a_e + 1$.
            \ElsIf{$\frac{a_e}{a_s} \geq 1$}
                \State $p_c^e=a_e/a_s - 1$.
            \EndIf
            \State $k_c^e = \frac{2}{\pi}\operatorname{arctan}(\lambda_c^e~p_c) + 1$,
            \State ${\boldsymbol{F}_{c}^{e}}^{'} = k_c^e \boldsymbol{F}_c^e$.
        \EndIf
    \State\Return ${\boldsymbol{F}_{c}^{e}}^{'}$
\end{algorithmic}
\end{algorithm}

Firstly, we input the major components of the current binding point F/T sensor~(such as $F_x^e$, $F_y^e$, and $F_z^e$) and the major components of the above binding point F/T sensor~(such as $F_x^s$, $F_y^s$, and $F_z^s$).
If either all MC of FA or all MC of UA are respectively smaller than their joint movement threshold ${\boldsymbol{F}_{mc}^e}_{\_th}$ and ${\boldsymbol{F}_{mc}^s}_{\_th}$, it represents that only the forearm or only the upper arm has movement intent. Therefore, it indicates that the users are in joint-oriented movement. 
When there are major components greater in both UA and FA than the threshold, it indicates movement in both the forearm and upper arm, suggesting the presence of target-oriented movement, with the forearm as the target. 
However, the level of coordination depends on the magnitude difference between UA and FA's two major components.
We calculate the magnitude of the MC at the elbow $a_e$ and shoulder $a_s$,
\begin{equation}
    a_e=\frac{\|{\boldsymbol{F}_{mc}^{e}}\|_2}{\|{{\boldsymbol{F}_{mc}^{e}}_{\_c}}\|_2}, ~a_s=\frac{\|{\boldsymbol{F}_{mc}^{s}}\|_2}{\|{{\boldsymbol{F}_{mc}^{s}}_{\_c}}\|_2},
\end{equation}
where~${\boldsymbol{F}_{mc}^{e}}_{\_c}$~and~${\boldsymbol{F}_{mc}^{e}}_{\_c}$~are coordination parameters of the coordination component~(CC) implemented to FCM of the shoulder and elbow joint respectively, to prevent abnormal values from appearing in $a_e$ and $a_s$. We compare them and obtain the ratio $p^e_c$,
\begin{equation}
    p_c^e=\left\{\begin{array}{cl}
        -a_s/a_e + 1, &~\textbf{if}~~{a_e}/{a_s} < 1 \\
         a_e/a_s - 1, &~\textbf{else}
    \end{array}\right.
\end{equation}
The ratio $p_c^e$ is input into \eqref{equ:k_c^e} to obtain the gain coefficient of the coordination component, whose numerical change is shown in~\figref{fig:curve}~(c). 
\begin{equation}
    k_c^e=\frac{2}{\pi}\operatorname{arctan}(\lambda_c^e~ p_c^e) + 1,
    \label{equ:k_c^e}
\end{equation}
where $\lambda_c^e$ denotes the adjustment parameter for the rate of change in gain. 
The influence of the coordinate components increases as the ratio of the MC of the forearm to the MC of the upper arm increases $p_c$. However, if $p_c$ is close to zero, this influence decreases to zero. Conversely, if $p_c$ increases significantly, the gain of this influence is gradually capped to maintain safety and stability.

For the wrist, the accompanying motion of WR1 is generated by $T^{w}_{x}$, while the accompanying motion of WR2 is generated by $F^{w}_{z}$ and WR3 is generated by $F^{w}_{y}$. 
When applying the coordination mechanism, the coordination components of the hand F/T sensor, $F^{w}_{z}$ and $F^{w}_{y}$, are input into IDM along with the major components of the upper arm and forearm F/T sensors. Similar to the coordination mechanism mentioned above, where the forearm serves as the target for target-oriented intent, here the major components of the hand~$\boldsymbol{F}_{mc}^{w}$ replace the major components at the forearm~$\boldsymbol{F}_{mc}^{e}$, and the sum of the major components of the upper arm and forearm sensors~$\boldsymbol{F}_{mc}^{e}+\boldsymbol{F}_{mc}^{s}$ replaces the major component of the upper arm~$\boldsymbol{F}_{mc}^{s}$. We calculate the magnitude of the MC at the wrist $a_w$ and the combination of the elbow and the shoulder $a_{es}$, 
\begin{equation}
    a_w=\frac{\|{\boldsymbol{F}_{mc}^{w}}\|_2}{\|{{\boldsymbol{F}_{mc}^{w}}_{\_c}}\|_2}, ~a_{es}=\frac{\|{\boldsymbol{F}_{mc}^{e}}\|_2}{\|{{\boldsymbol{F}_{mc}^{e}}_{\_c}}\|_2}+\frac{\|{\boldsymbol{F}_{mc}^{s}}\|_2}{\|{{\boldsymbol{F}_{mc}^{s}}_{\_c}}\|_2}.
\end{equation}
We compare them and obtain the ratio $p_c^w$,
\begin{equation}
    p_c^w=\left\{\begin{array}{cl}
        -a_{es}/a_w + 1, &~\textbf{if}~~{a_w}/{a_{es}} < 1\\
         a_w/a_{es} - 1, &~\textbf{else}
    \end{array}\right.
\end{equation}

\begin{equation}
    k_c^w=\frac{2}{\pi}\operatorname{arctan}(\lambda_c^w~ p_c^w) + 1.
    \label{equ:k_c^w}
\end{equation}
The coordination components in elbow~$\boldsymbol{F}_c^e$~and the coordination components in wrist~$\boldsymbol{F}_c^w$ are adjusted to obtain:
\begin{equation} 
{\boldsymbol{F}_{c}^{e}}^{'} = k_c^e \boldsymbol{F}_c^e,~~{\boldsymbol{F}_{c}^{w}}^{'} = k_c^w \boldsymbol{F}_c^w.
\end{equation}
Finally, through the intention distinction module, the coordinated control torques output by FCM for each joint are
\begin{equation}                    
       \boldsymbol{\tau}\!_{_{F\!C\!M}}^{{\,}s}={\boldsymbol{J}\!_{\text{\tiny{B}}}^{{\,}\text{\tiny{UA}}}}^{\!\top}{\boldsymbol{F}_{c}^{s}}^{'},~~
       \boldsymbol{\tau}\!_{_{F\!C\!M}}^{{\,}e}={\boldsymbol{J}\!_{\text{\tiny{SC1}}}^{{\,}\text{\tiny{FA}}}\!^\top}{\boldsymbol{F}_{c}^{e}}^{'},~~
       \boldsymbol{\tau}\!_{_{F\!C\!M}}^{{\,}w}={\boldsymbol{J}\!_{\text{\tiny{SC1}}}^{{\,}\text{\tiny{HA}}}\!^\top}{\boldsymbol{F}_{c}^{w}}^{'}.
    \label{equ:tau_a_s}
\end{equation}
\begin{equation}
\boldsymbol{\tau}\!_{_{F\!C\!M}}=\left[\begin{array}{l}
\boldsymbol{E}_{4 \times 4} \\
\boldsymbol{O}_{5 \times 4}
\end{array}\right] \boldsymbol{\tau}\!_{_{F\!C\!M}}^{{\,}s}+\left[\begin{array}{l}
\boldsymbol{O}_{1 \times 5} \\
\boldsymbol{E}_{5 \times 5} \\
\boldsymbol{O}_{3 \times 5}
\end{array}\right] \boldsymbol{\tau}\!_{_{F\!C\!M}}^{{\,}e}+\left[\begin{array}{c}
\boldsymbol{O}_{1 \times 8} \\
\boldsymbol{E}_{8 \times 8}
\end{array}\right] \boldsymbol{\tau}\!_{_{F\!C\!M}}^{{\,}w}
\label{equ:tau_fcm}
\end{equation}

\subsection{Dynamic feedforward compensation}
We use dynamics compensation as the feed-forward to the control framework, as close as possible to the actual model, which contains inertia, Coriolis force, gravity, and friction compensation. Among them, gravity and friction are becoming more important parts.
The gravity compensation amount is determined by the statics model and adjusted through gravity balance experiments. We measured the mass and center of mass of each exoskeleton link using measurement methods.
Friction compensation involves the use of the parameter identification method to identify the friction model of each joint. 
The static and dynamic friction forces of the joint motor are critical parameters. 
Due to the weak reverse driveability of the harmonic reduction motor we use, the static friction force is typically greater than the dynamic one. 
When the motor starts and when the direction of movement changes (which is common in rapid arm movements), the drive torque used as friction compensation will experience a sudden change,  which leads to vibrations in the exoskeleton.
Therefore, we use a combination of static friction, Coulomb viscous friction, and Stribeck friction models to model the friction of each motor.
Due to the presence of noise in the motor data feedback and the large range of discontinuous driving force jumps around the speed of 0, which is not conducive to the static stability of the operating arm, it is also possible to cause vibration.
Therefore, we utilize a smoothing function to transition these sudden changes in compensation forces:
\begin{equation}
\hat{\boldsymbol{f}}=\frac{2 \boldsymbol{f_c}}{\boldsymbol{\pi}}\operatorname{arctan}(\frac{\boldsymbol{v}}{\boldsymbol{a}})+(\boldsymbol{f_s}-\boldsymbol{f_c})e^{-\lvert{\frac{\boldsymbol{v}}{\boldsymbol{v_s}}\lvert}} \text{sgn}(\boldsymbol{v}),
\end{equation}
where $\boldsymbol{f_c}$ and $\boldsymbol{f_s}$ represent the critical values of dynamic and static friction. 
They can be obtained by measuring the starting torque and the constant speed torque of each joint without being affected by gravity. $\boldsymbol{v_s}$ is the velocity parameter of the Striibeck friction model, used to adjust the decay effect of static friction that transitions to kinetic friction. The supplementary material~(D) includes the specific parameter values.

For the compensation of inertia and Coriolis forces, we use a parameter identification method to obtain the moments of inertia for each link. Subsequently, based on the formulas for dynamic modeling, we calculate the compensation amounts for inertia and Coriolis forces. The relevant parameters are included in the supplementary material~(D).

%%%%%%%%%%%%%%%%%%%%%%%%%%%%%%%%%%%%%%%%%%%%%%%%%%%%%%%%%%%%%%%%%%%%%%%%%%%%%%%%
\section{Experiments and Results}
\label{sec:exp}

Researches on exoskeletons mainly focus on improving individual performance and conducting experimental evaluations on some indicators without proposing a comprehensive assessment method to evaluate the various aspects of exoskeleton performance~\cite{Huang2019tsmc}. Therefore, five sets of experiments were conducted to evaluate the performance of the exoskeleton and our control strategy from five aspects: \textbf{flexibility}, \textbf{adaptability}, \textbf{accuracy}, \textbf{speed}, and \textbf{fatigue} level.
The experiments can serve as a comprehensive example of experimental evaluation for upper limb exoskeletons.

\begin{table}[tbp]
    \centering
    \caption{Volunteers physical information}
    \begin{tabular}{clcccc}
    \toprule
         \# & Gender&  Height &  Weight&  Upper Arm & Forearm\\
    \midrule
         1& Male&  175\,cm& 65\,kg & 318mm & 261mm\\
         2& Male&  177\,cm& 75\,kg & 323mm & 267mm \\
         3& Female& 165\,cm& 51\,kg& 293mm & 255mm \\
         4& Female& 163\,cm& 48\,kg& 290mm & 251mm \\
         5& Male&  180\,cm& 80\,kg & 332mm & 273mm\\
    \bottomrule
    \end{tabular}
    \label{tab:volunteer}
      \vspace{-0.6cm}
\end{table}

\subsection{Flexibility}

\begin{figure*}[tbp]
    \centering
    %Table1%
\vspace{-0.7cm}
    \begin{minipage}{0.9\linewidth}
            \includegraphics[width=1\linewidth]{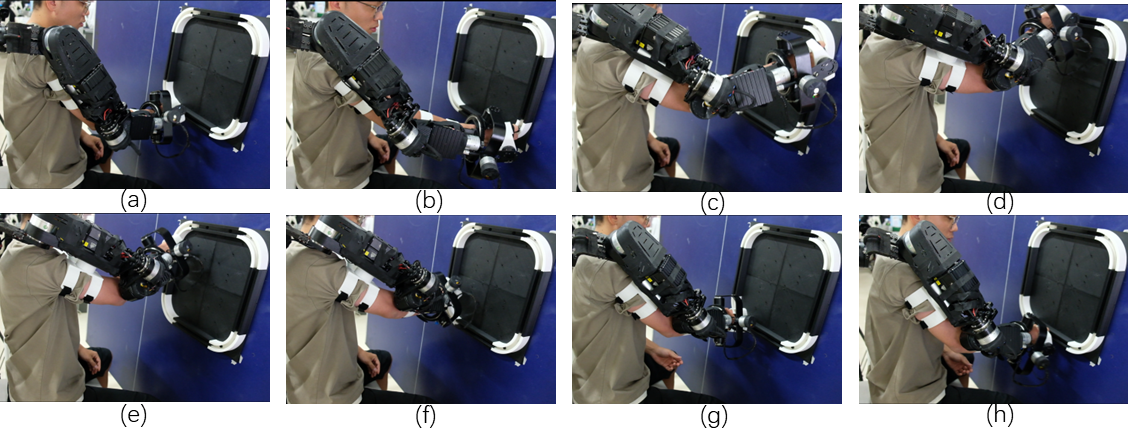}
            \vspace{-0.8cm}
            \caption{Moving the slider along the track while wearing the exoskeleton.}
            %\caption{Motivating Example. }
            \label{fig:guidao}
    \end{minipage}
    \vspace{-0.1cm}
    \begin{minipage}{0.32\linewidth}
            \includegraphics[width=1\linewidth]{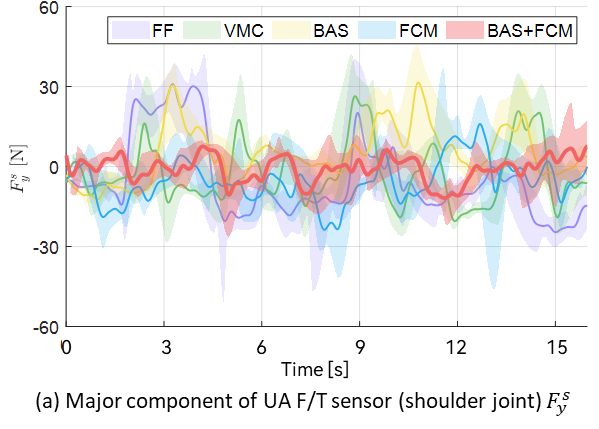}
            %\caption{Motivating Example. }
            %\label{fig:1-1}
    \end{minipage}    
    \begin{minipage}{0.32\linewidth}
            \includegraphics[width=1\linewidth]{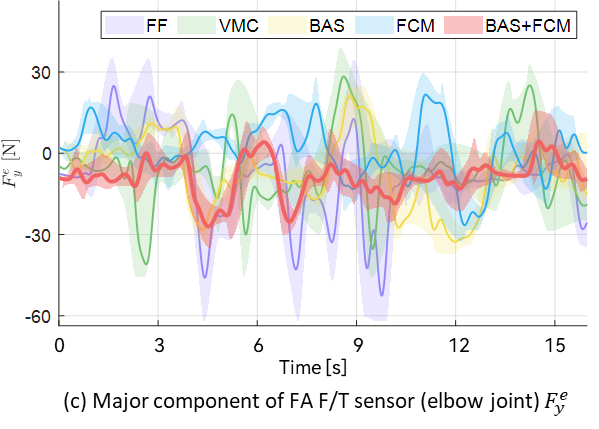}
            %\caption{Motivating Example. }
            %\label{fig:1-2}
    \end{minipage}   
    \begin{minipage}{0.32\linewidth}
            \includegraphics[width=1\linewidth]{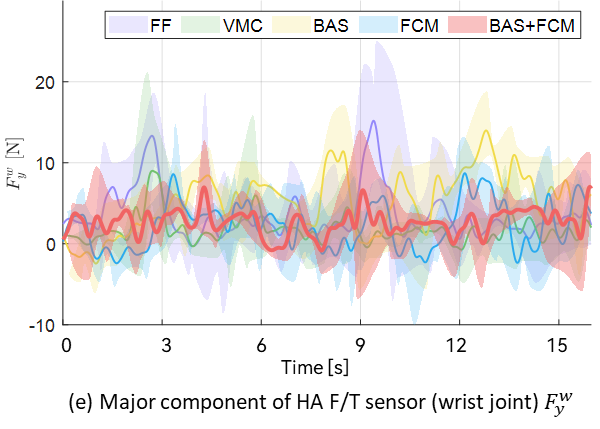}
            %\caption{Motivating Example. }
            %\label{fig:1-3}
    \end{minipage}   
    \begin{minipage}{0.32\linewidth}
            \includegraphics[width=1\linewidth]{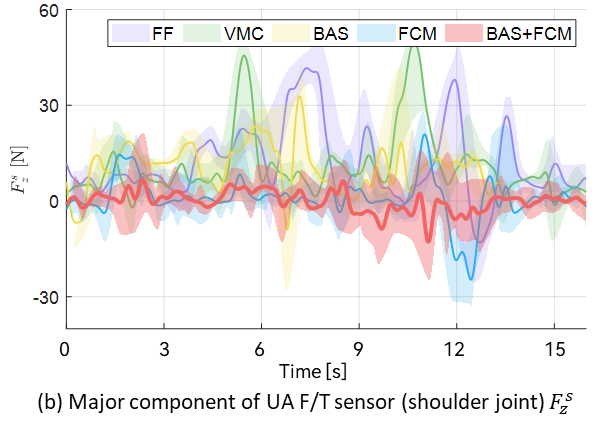}
            %\caption{Motivating Example. }
            %\label{fig:2-1}
    \end{minipage} 
    \begin{minipage}{0.32\linewidth}
            \includegraphics[width=1\linewidth]{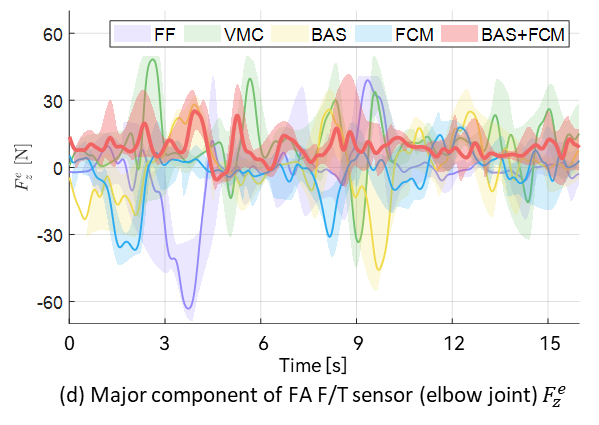}
            %\caption{Motivating Example. }
            %\label{fig:2-2}
    \end{minipage} 
    \begin{minipage}{0.32\linewidth}
            \includegraphics[width=1\linewidth]{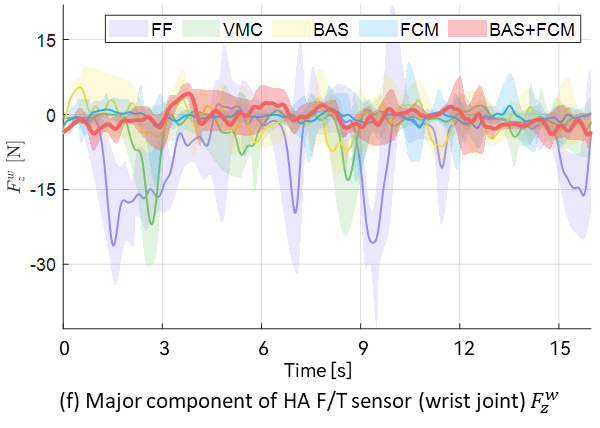}
            %\caption{Motivating Example. }
            %\label{fig:2-3}
    \end{minipage} 
    
    %Table 1%
    \begin{minipage}{0.64\linewidth}
    \renewcommand\arraystretch{1}
    \setlength{\tabcolsep}{2pt}
    \footnotesize
    \centering
    \vspace{-0.5cm}
    \captionof{table}{Average absolute values~(MAV) and absolute value standard deviations~(MAD) of each major component using different methods.~(MAV\,/\,MAD)}
    \begin{tabular}{cccccccc}
      \toprule
        Method & $F^s_y$ & $F^s_z$ & $F^e_y$ & $F^e_z$ & $F^w_y$  & $F^w_z$  & $T^w_x$ \\
        \midrule
        FF & 13.44\,/\,11.27 & 13.19\,/\,9.44 & 13.53\,/\,10.49 & 8.11\,/\,8.21 & 3.60\,/\,2.12 & 6.03\,/\,5.86 &  0.10\,/\,0.07 \\ 
        VMC & 9.59\,/\,8.869 & 12.289\,/\,7.33 & 10.939\,/\,8.35 & 12.32\,/\,8.02 & 1.82\,/\,1.11 & 2.35\,/\,2.38 &  0.10\,/\,0.096 \\ 
        BAS & 8.44\,/\,7.99 & 11.11\,/\,5.44 & 11.11\,/\,9.72 & 10.33\,/\,10.28 & 5.36\,/\,2.95 & 2.30\,/\,2.14 & 0.29\,/\,0.17\\ 
        FCM & 7.89\,/\,5.91 & 4.12\,/\,4.11 & 7.57\,/\,7.44 & 6.89\,/\,7.13 & 2.88\,/\,2.41 & 1.76\,/\,0.63 &  0.096\,/\,0.084\\
        BAS+FCM & \textbf{3.62}\,/\,\textbf{3.64} & \textbf{2.63}\,/\,\textbf{2.64} & \textbf{5.34}\,/\,\textbf{4.73} & \textbf{4.98}\,/\,\textbf{3.46} & \textbf{2.70}\,/\,\textbf{1.14} & \textbf{1.56}\,/\,\textbf{1.41} & \textbf{0.083}\,/\,\textbf{0.071}\\ 
        \bottomrule
    \end{tabular}
    \label{tab:flexibility}
    \vspace{-0.3cm}
    \end{minipage}
    \hspace{0.2cm}
    \begin{minipage}{0.32\linewidth}
            \includegraphics[width=1\linewidth]{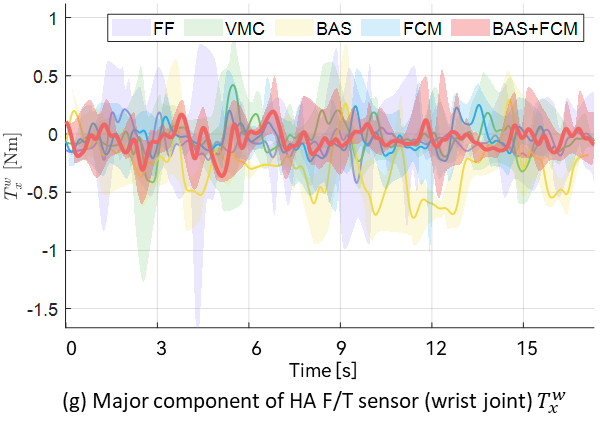}
            %\caption{Motivating Example. }
    \vspace{-0.3cm}
    \end{minipage} 
    \vspace{-0.4cm}
    \captionof{figure}{The average curve of the major component~(MC) of interaction force measured during slider movement tests using an exoskeleton equipped with five different control methods. The shaded area indicates the range of deviation. The results of the FF algorithm are indicated in purple, VMC results in green, BAS results in yellow, FCM results in blue, and the combined algorithm of BAS+FCM in red.}
    \label{fig:fexibility}
  \vspace{-0.5cm}
\end{figure*}

\begin{figure*}[tbp]
    \vspace{-0.6cm}
    \centering
    \includegraphics[width=1\linewidth]{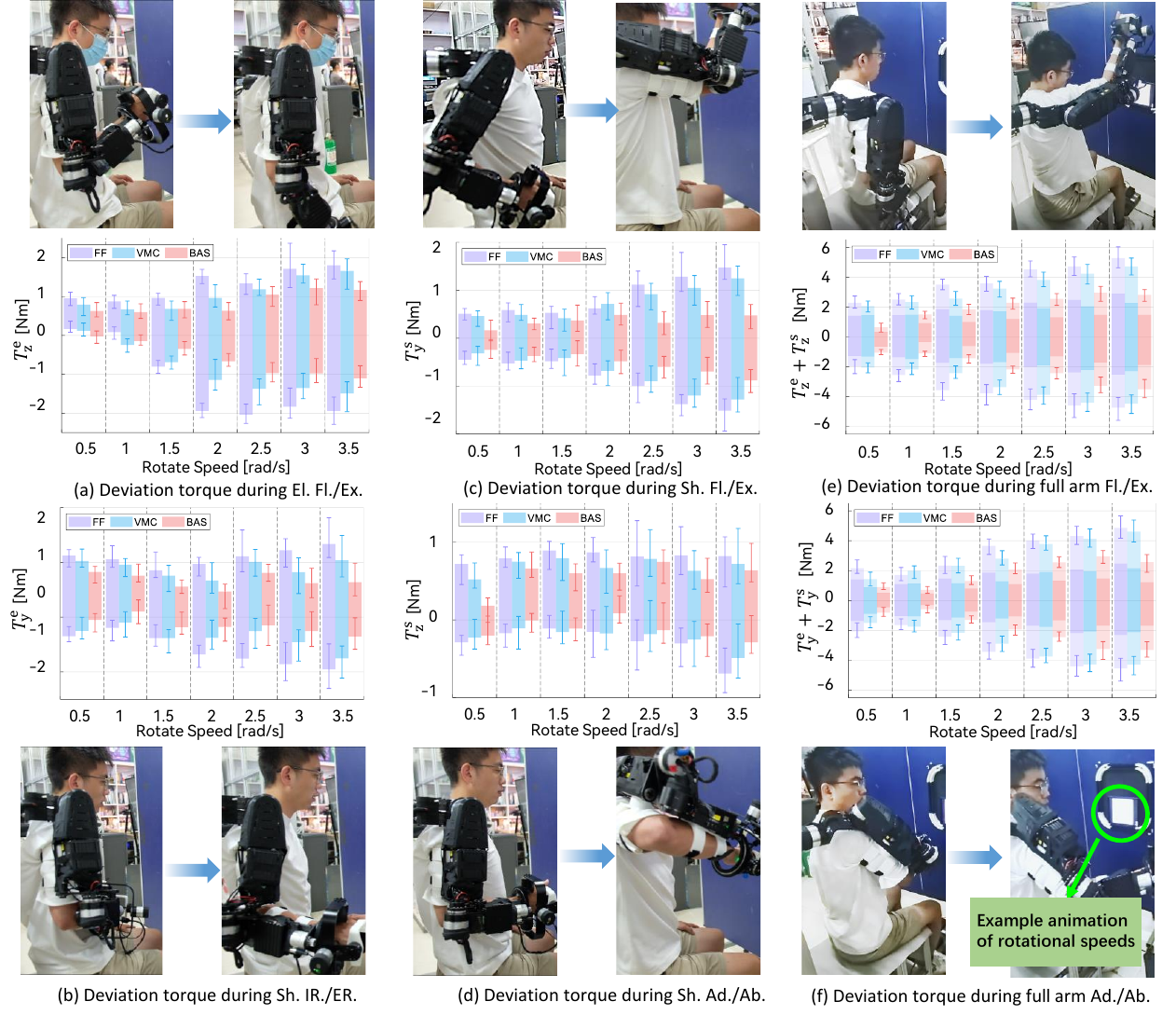}
    \vspace{-0.6cm}
    \caption{(a), (b), (c), and (d) show volunteers performing the following single-joint movements: Elbow Flexion/Extension~(El.\,Fl./Ex.), Shoulder Flexion/Extension~(Sh.\,Fl./Ex.), Shoulder Internal/External Rotation~(Sh\,IR./ER.), and Shoulder Adduction/Abduction~(Sh.\,Ad./Ab.). The changes in the assistant component torque $T_z^{e}$, $T_y^{e}$, $T_z^{s}$, $T_y^{s}$ were measured at different speeds for each movement. The bar chart shows the average maximum disturbance for all volunteers during exercise, with error bars indicating the maximum deviation of individual samples. (e) and (f) show the variation in maximum disturbance levels at different speeds during two compound movements: the full arm lift (El.\,Fl./Ex. with Sh.\,Fl./Ex.) and the full arm outward swing~(Sh.\,IR./ER. with Sh.\,Ad./Ab.). The bar charts represent the sum of the assistant component of two joint motions, $T_z^{e}+T_z^{s}$, $T_y^{e}+T_y^{s}$, in different speeds. The dark area of the bar chart represents $T_z^{e}$ and $T_y^{e}$, while the light area represents $T_z^{s}$ and $T_z^{s}$. The three colors correspond to the three algorithms: FF, VMC, and Ours.}
    \label{fig:guidao_curve}
  \vspace{-0.5cm}
\end{figure*}

\textbf{Large range and high-speed motion along a fixed track.}
  \vspace{-0.1cm}
            Five volunteers move a slider along the square track while wearing the exoskeleton and using different control methods, as shown in~\figref{fig:guidao}. The volunteers were provided with written informed consent to participate in the experiment in accordance with the Declaration of Helsinki. Physical information of the volunteers is shown as \tabref{tab:volunteer}.
            The track is a square with a side length of 40 cm, and the corners are rounded with a radius of 10 cm. Before the experiment, the operators underwent a 10-minute training session to achieve the goal of completing a complete circuit of the track in approximately 20 seconds. To maintain the generality of the data, each operator conducted five tests under five different control algorithms: The feed-forward compensation with the traditional decoupling active force controller~(FF)~\cite{Fabian2018Automatisierungstechnik}, the virtual mass control~(VMC)~\cite{zimmermann2020iros}, feed-forward compensation only with donning disturbance handling~(FF + BAS), feed-forward compensation only with whole-arm coordination mechanism~(FF+FCM), and combination of two methods (BAS+FCM), resulting in a total of 125 exercise tests~(5\,$\times$\,5\,$\times$\,5). The interaction forces are measured using F/T sensors, and \figref{fig:fexibility} shows the numerical curves of the major force components with the different algorithms implemented throughout the motion. The shaded areas represent the range of values from 25 different data distributions under the same algorithm, while the solid lines represent the averages. 
            \tabref{tab:flexibility} displays the average absolute values~(MAV) and the absolute value standard deviations~(MAD) of each major component of interactive forces in the shoulder, elbow and wrist by different methods.

            The experimental data show that when applying the FF benchmark algorithm, the interaction force in the shoulder is approximately 13\,N; at the elbow, the interaction force is 10\,N. The VMC method significantly improves transparency, requiring less interaction force to drive similar movements. Looking at the individual use of BAS and FCM, FCM offers a higher enhancement in transparency, while the use of BAS alone provides limited improvements. It aligns with our initial design intent, as BAS primarily addresses disturbances at the binding points, and the sliding motion of the slider is more directed toward the intended movement of the arm's endpoint. Therefore, FCM can greatly enhance flexibility. When both methods are used together, the interaction force in the shoulder is approximately 3\,N, and at the elbow, it is about 5\,N, which substantially promotes the improvement of transparency. The interaction forces decreased by 73\,\%, 82\,\%, 61\,\%, 39\,\%, 25\,\%, 74\,\%, and 17\,\% respectively, particularly at the shoulder and elbow, where the reduction of the interaction force is significant. The standard deviation of the data has similarly decreased. The results validate that our control method can effectively enhance flexibility during rapid movements.
\vspace{-0.3cm}            
\subsection{Adaptability}
\vspace{-0.2cm}  
\textbf{Interactive force disturbances tests in different speeds.}

To demonstrate the effectiveness of the adaptability of our BAS mechanism, five volunteers continued to perform four sets of single joint movements and two sets of compound joint movements using the three algorithms (FF, VMC, and BAS) implemented in the exoskeleton as shown in \figref{fig:guidao_curve}. 

When each volunteer performs joint movements, they will have a screen in front of them that loops example animations, as indicated in \figref{fig:guidao_curve}~(f). The example animations consist of adjustable fixed-speed back-and-forth or up-and-down pendulum motions. Volunteers wearing the exoskeleton must ensure that their motion speed matches the pendulum motion speed set in the animations. 
Here, we tested six groups of upper limb movements. Each group of movements with speed increased from 0.5\,rad/s, incrementally by 0.5\,rad/s, to 3.5\,rad/s (resulting in a total of 630 movements: 5 participants\,$\times$\,3 algorithms\,$\times$\,6 speeds\,$\times$\,7 movements). For each speed, each movement involved the joint moving back and forth 5 times.
(a), (b), (c) and (d) are performing the following single-joint movements: Elbow Flexion/Extension~(EL.\,Fl./Ex.), Shoulder Flexion/Extension~(Sh.\,Fl./Ex.), Shoulder Internal/External Rotation~(Sh.\,IR./ER.), and Shoulder Adduction/Abduction~(Sh.\,Ad./Ab.). (c), (d) involve performing the two compound movements: the full arm lift (EL.\,Fl./Ex. with Sh.\,Fl./Ex.) and the full arm outward swing~(Sh.\,IR./ER. with Sh.\,Ad./Ab.).
We then measured the assistant component~(AC) of the interaction forces. The magnitude of AC represents the interactive force caused by 
the donning deviations because, in single-joint or compound joint movements, the presence of numerical values in AC indicates that the exoskeleton does not fit perfectly with the human arm, which signifies the existence of interaction disturbances. The statistical values of AC measured from the force sensors for each algorithm under different speed standards are depicted in \figref{fig:guidao_curve}.

It is evident from the bar charts that as the motion speed increases, the values of AC data also increase. However, from an overall perspective of the curves, when implementing the BAS method we proposed, the AC values are significantly lower than that of the curves implementing other methods. In addition, with this method, the increase in AC values with speed is also relatively smaller than that seen with other methods. In addition, with this method, the increase in AC values with speed is also relatively smaller than that seen with other methods.
In compound movements involving two joints, the sum of the assistant components is greater than the sum of the movements of the two joints individually, as indicated by the data, which shows that more donning deviations are generated in such compound movements. 
However, our proposed BAS mechanism still reduces the disturbance level to a relatively low level, demonstrating its effectiveness in suppressing disturbances during high-speed movements, especially under compound movements. For a more detailed discussion of the numerical results, refer to the supplementary material~(E).

\begin{figure}[tbp]
    \begin{minipage}{0.9\linewidth}
                \includegraphics[width=1\linewidth]{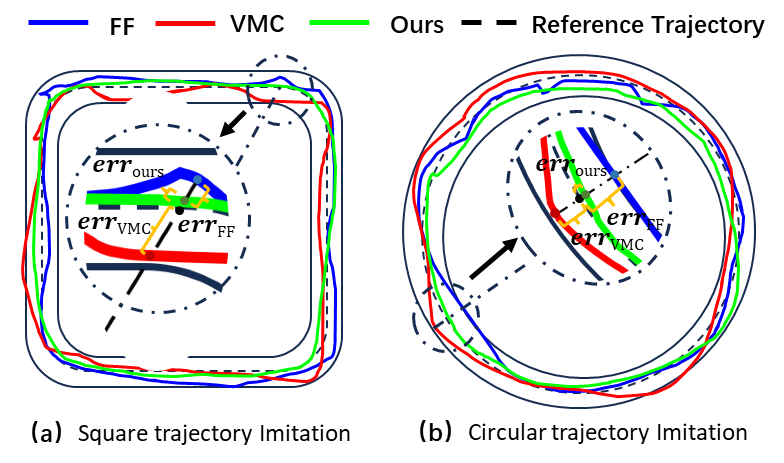}
    \end{minipage}  
    \centering
    \begin{minipage}{0.9\linewidth}
        \includegraphics[width=1\linewidth]{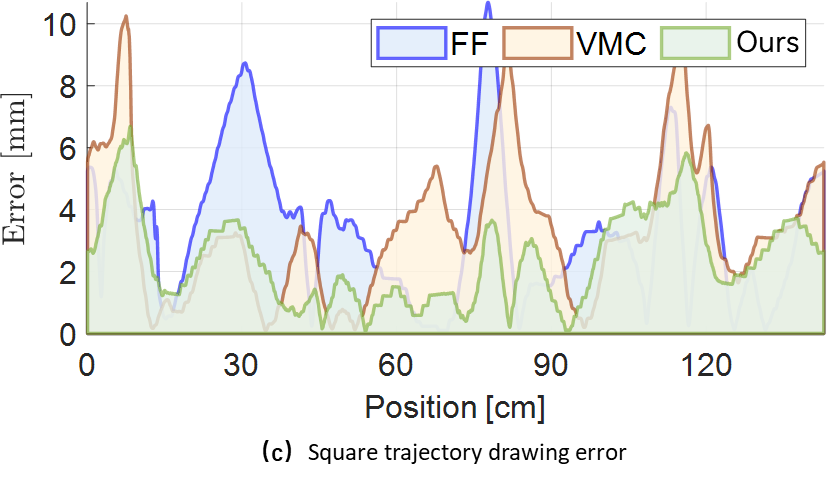}
        \includegraphics[width=1\linewidth]{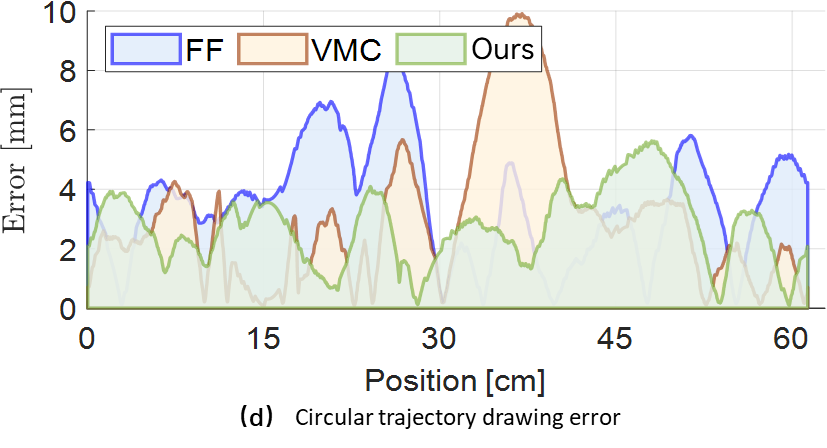}
        %\caption{Motivating Example. }
    \caption{Error comparison in drawing trajectory with the exoskeleton using the three control methods}
    \label{fig:guiji}
    \end{minipage}
    \vspace{0.2cm}
    \begin{minipage}{1\linewidth}
    \vspace{0.2cm}
    \captionof{table}{Mean and standard deviation in accuracy experiments}
    \begin{tabular}{c|ccc|ccc}
      \toprule
        \multirow{2}*{Method} & \multicolumn{3}{c|}{Square orbit} &\multicolumn{3}{c}{Circular orbit} \\
        \cline{2-7}
        ~ & FF & VMC & Ours & FF & VMC & Ours\\
        \midrule
        Mean & 3.314 & 3.600 & \textbf{2.401} & 3.539 & 3.011 & \textbf{2.733}  \\
        Std. & 2.267 & 2.260 & \textbf{1.445} & 1.933 & 2.399 & \textbf{1.253} \\ 
        \bottomrule
    \end{tabular}
    \label{tab:guiji}
    \vspace{-0.4cm}
    \end{minipage}
    \vspace{-0.4cm}
\end{figure}
\vspace{-0.3cm}
\subsection{Accuracy}   
\textbf{Error in quickly tracing the established trajectory}.
To verify the rapid operational precision of exoskeletons equipped with different control algorithms, operators must wear the exoskeleton and trace a square and a circular trajectory, as shown in \figref{fig:guiji}. 
An A2 sheet paper containing the reference trajectory and boundary is posted on the board before the operators.
The operators, wearing an exoskeleton, hold a pencil to quickly trace a circle along the reference trajectory within 15 seconds~(the bottom left corner of the trajectory is the starting point). The drawing should adhere closely to the reference curve. 
The dotted line represents the reference trajectory, while the solid lines on both sides serve as visual constraints for the operator. The blue, red, and green curves in the figure represent the trajectories drawn using the FF, VMC, and our algorithm, respectively. 
The experimental trajectory paper will be scanned and binarized to extract the trajectory of each implemented method. Then, radial lines will be drawn from the center of the reference trajectory.
The intersection points of the radial line and the drawn trajectories, along with the intersection point of the radial line and the reference trajectory, represent the drawing errors of these algorithms, as shown in the labeled enlarged images in \figref{fig:guiji}~(a)~and~(b). Therefore, the horizontal axis labeled "position" in~(c)~and~(d)~indicate the positions moving counterclockwise from the starting point to the endpoint of the reference trajectory.
After applying our control method, the average errors for the two trajectories were 2.401 mm and 2.733 mm, with standard deviations of 1.445 and 1.253, respectively, which are the smallest compared to other methods and represent over a 30\,\% improvement. In addition, it can be observed from the trajectory and curve graphs that the area and peak values of the green area are significantly smaller than those of the blue and red areas, which demonstrates that the accuracy and stability of movements using our algorithm are comparatively better.  
\vspace{-0.2cm}
\begin{figure}[tbp]
    \centering
    \begin{minipage}{1\linewidth}
        \includegraphics[width=1\linewidth]{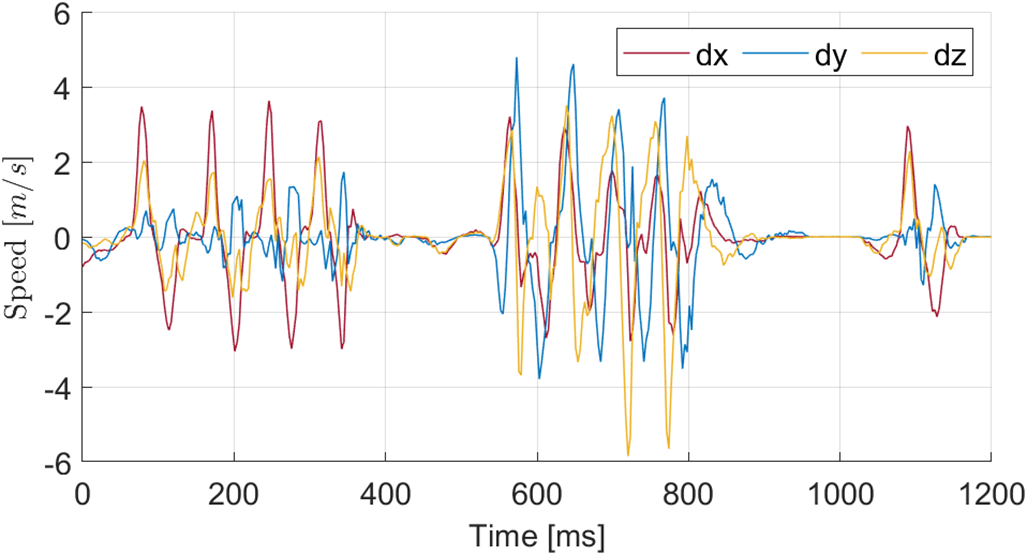}
        %\caption{Motivating Example. }
    \caption{Linear speed of the end of the exoskeleton during the fast dynamic motion.}
    \label{fig:speed_end}
    \end{minipage}
    \begin{minipage}{1\linewidth}
    \vspace{0.3cm}
    \captionof{table}{Each joint Max speed~(Unit:\,rad/s), max linear~(Unit:\,m/s), and angular speed~(Unit:\,rad/s) during fast dynamic motion with the exoskeleton.}
    \vspace{-0.3cm}
    \renewcommand\arraystretch{1.2}
    \setlength{\tabcolsep}{3pt}
    \centering
    \begin{tabular}{cccccccc}
      \toprule
        $\dot{\theta}_2$ & $\dot{\theta}_3$ & $\dot{\theta}_4$ & $\dot{\theta}_5$ & $\dot{\theta}_6$ & $\dot{\theta}_7$ & $\dot{\theta}_8$ & $\dot{\theta}_9$ \\
        \midrule
        3.251 & 2.444 & 6.631 & \textbf{9.319} & 8.152 & 4.364 & 3.521 & 2.973 \\
        \cline{1-8}
        \multicolumn{3}{c}{Max linear speed:} & \textbf{5.721} ~{|} & \multicolumn{3}{c}{Max angular speed:} & \textbf{9.372}\\
        \bottomrule
    \end{tabular}
    \label{tab:MaxSpeed}
    \end{minipage}    
    \vspace{-0.4cm}
\end{figure}
\vspace{-0.2cm}
\subsection{Speed} 

\textbf{Maximum Speed Test.}
Volunteer {\#1} wearing the exoskeleton performed a quick punching action and a rapid circular motion in the air, recording the end-effector speed in three linear directions and the joint speeds of the exoskeleton, as shown in \figref{fig:speed_end}. The other three angular directions and joint speeds are presented in the supplementary material~(F). The maximum speed of each joint, the maximum linear speed in the end direction, and the maximum rotational speed are shown in \tabref{tab:MaxSpeed}; the experimental data show that our proposed BAS + FCM method deployed in the designed exoskeleton achieves a maximum joint speed of 9.319\,rad/s, a maximum end-effector speed of 5.721\,m/s in the linear direction, and a maximum end-effector speed of 9.372\,rad/s in the rotational direction. It is the maximum operating speed reported for the full-arm exoskeleton\cite{zimmermann2023tro}, which is close to the limit of a typical human punching speed. Our results outperform the baseline methods in all the metrics reported for exoskeletons, which validate the advancement of our algorithm in high-dynamic situations.

\begin{figure}[tbp]
    \centering
    \includegraphics[width=0.9\linewidth]{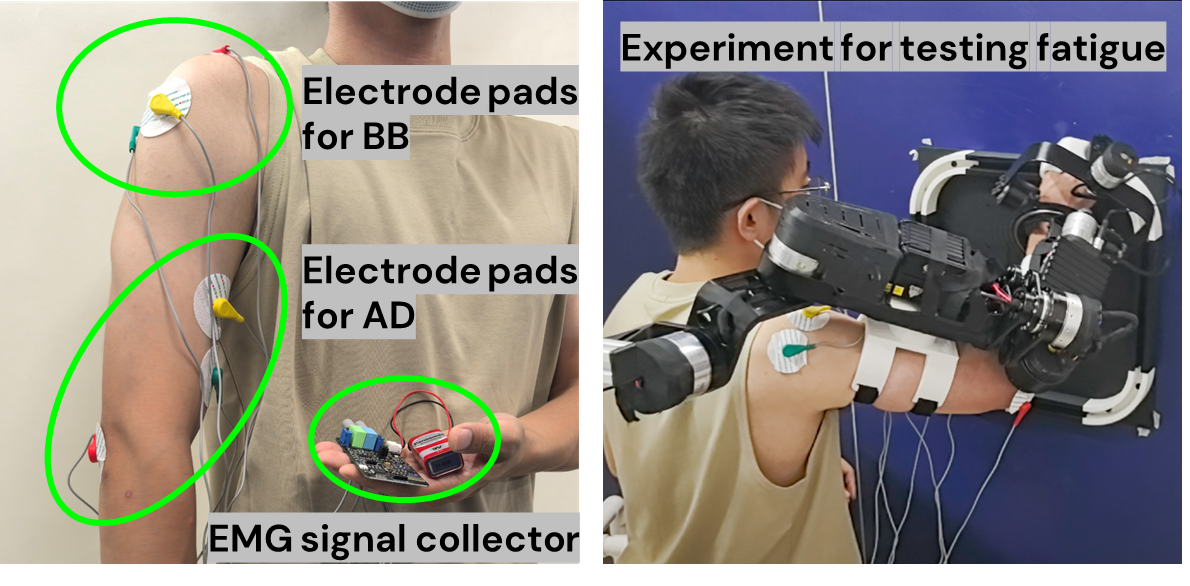}
    \caption{Using EMG to test the fatigue level of the exoskeleton.}
    \label{fig:jidian}
    \vspace{-0.5cm}
\end{figure}

\begin{figure}[tbp]
    \centering
    \includegraphics[width=1.02\linewidth]{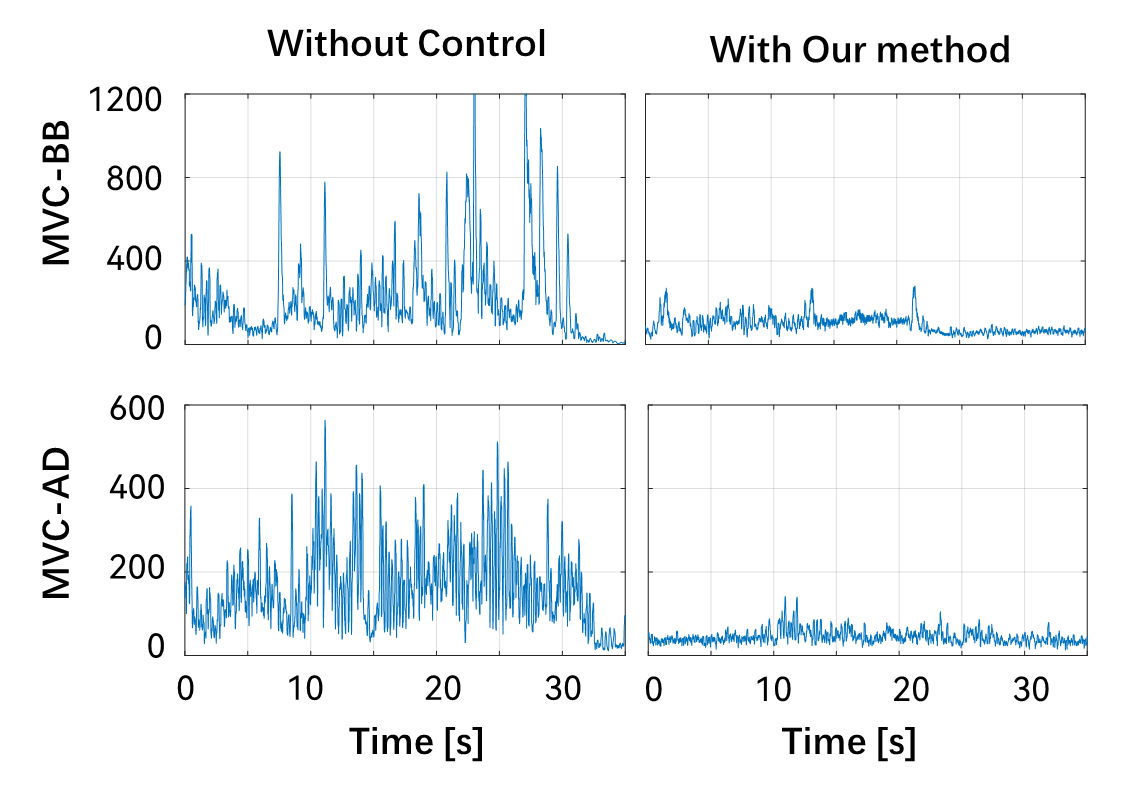}
    \vspace{-0.7cm}
    \caption{Muscular activation of voluntary contraction of biceps brachii~(BB) and anterior deltoid~(AD) while wearing the exoskeleton without control and with our control method. Other results of the experiment of Fatigue with EMG}
    \label{fig:MVC2}
    \vspace{-0.6cm}
\end{figure}
\vspace{-0.3cm}
\subsection{Fatigue}
\vspace{-0.1cm}
\textbf{Measuring the muscle voluntary contraction~(MVC) using electromyography~(EMG).}
    Based on the first set of the flexibility experiment, we use electromyography equipment~(EMG) to measure muscle activation of biceps brachii~(BB) and anterior deltoi~(AD) muscles\cite{Kim2020rcim}. These two muscle groups are frequently involved during daily activities of the arm. Our measurement method is illustrated in \figref{fig:jidian}. Two muscles are connected to two sets of electrode pads, each set consisting of three electrode patches. The yellow electrode is placed where the greatest muscle deformation is measured. The green electrode serves as a reference electrode and is placed near the yellow electrode. In contrast, the red electrode acts as a baseline electrode and is positioned at the connection between the muscle and the bone, where muscle deformation is the least.  
    EMG signal collector measured the muscle voluntary contraction (MVC) of the target muscles with wearing the exoskeleton under different control methods~(FF and our method) and without the exoskeleton. The VMC method can adjust the level of power assistance; therefore, we did not set a comparison with it.
    
    From~\figref{fig:MVC2} and the supplementary material~(G), regarding the MVC of BB, the maximum data value slightly exceeded 300 when the volunteer did not wear the exoskeleton. When wearing the exoskeleton implemented in our method, the maximum data value is less than 300, and the overall magnitude has not changed much. The maximum data value is over 300 when the user wears the exoskeleton with the FF method, and the magnitude increases from 1300 to 2200 compared to without the exoskeleton. Regarding MVC of AD, the advantages of our method are even more apparent, the maximum value of the curve and the overall amplitude with exoskeleton using our method are lower than that of using the FF method and even that of without wearing exoskeleton. However, when using conventional FF control combined with decoupling active force control, it is difficult to achieve perfect dynamic compensation and coordinated compliant control. As a result, the same movements increase the maximum and average muscle loads. As shown in \tabref{tab:jidian}, the maximum value and mean for the exoskeleton with our method are 280, 85.33 for BB and 141, 45.09 for AD, which are all lower than without the exoskeleton and wearing it with the FF method. This indicates that using the exoskeleton implemented in our method does not increase operator fatigue.
    
\begin{table}[t]
    \centering
    \captionof{table}{The max value, mean, and standard deviation~(Std.) of the muscular activation of voluntary contraction of biceps brachii and anterior deltoid. W means with, and WO means without.}
    \footnotesize
    \begin{tabular}{cccc|ccc}
      \toprule
        ~ &\multicolumn{3}{c}{Biceps brachii~(BB)} & \multicolumn{3}{c}{Anterior deltoid (AD)} \\
        \midrule
        Method  & Max & Mean & Std. & Max & Mean & Std. \\
        \cline{1-7}
        WO Control & 1491 & 215.56 & 201.18 & 564.75 & 163.16 & 93.61\\
        W Exo~(FF) & 347 & 110.88 & \textbf{40.85} & 185 & 52.01 & 21.99\\
        WO Exo & 312 & 95.75 & 41.43 & 160 & 52.90 & 20.85\\
        W Exo~(ours) & \textbf{280} & \textbf{85.53} & 41.20 & \textbf{141} & \textbf{45.09} & \textbf{15.84}\\
        \bottomrule
    \end{tabular}
    \label{tab:jidian}
    \vspace{-0.5cm}
\end{table}
%%%%%%%%%%%%%%%%%%%%%%%%%%%%%%%%%%%%%%%%%%%%%%%%%%%%%%%%%
\vspace{-0.2cm}
\section{Conclusion}
\vspace{-0.1cm}
\label{sec:conclusion}
Through the analysis of the interactive force caused by the donning deviations, we explained that binding offsets are common and can negatively impact the control effectiveness of exoskeletons and that focusing on the coordination of the entire arm is vital.
Based on this, we propose the binding alignment strategy~(BAS) method to reduce the impact of these disturbances. Furthermore, the full-arm coordination mechanism~(FCM) is merged with BAS to construct a control system for the full-arm exoskeleton based on the classification of force sensor data. In addition, an intention distinction model is proposed to distinguish motion intentions and resolve the conflict between different force components. This system aims to reduce the discomfort of wearing and improve the coordinated flexibility of the exoskeleton in tandem with human arm movements. 
Finally, a comprehensive set of experiments tests the effects of these two proposed methods and their combination, which collectively can reduce the number of donning deviations and enhance coordination. The excellent accompanying control performance based on our method is beneficial for applying the exoskeleton in manipulation and rehabilitation scenarios.

In future research, based on this study, we will continue to research the force feedback control for exoskeletons under high dynamics. This will provide a research foundation for human-machine skill transfer between humans and robots.

%%%%%%%%%%%%%%%%%%%%%%%%%%%%%%%%%%%%%%%%%%%%%%%%%%%%%%%%%%%%%%%%%%%%%%%%%%%%%%%%
\bibliographystyle{ieeetr}

\bibliography{arxiv2024Cheng}
%%%%%%%%%%%%%%%%%%%%%%%%%%%%%%%%%%%%%%%%%%%%%%%%%%%%%%%%%%%%%%%%%%%%%%%%%%%%%%%%
\end{document}